\documentclass{article}

\PassOptionsToPackage{numbers, compress}{natbib}
\usepackage[final]{neurips_2023}

\usepackage[utf8]{inputenc} 
\usepackage[T1]{fontenc}    
\usepackage{hyperref}       
\usepackage{url}            
\usepackage{booktabs}       
\usepackage{amsfonts}       
\usepackage{nicefrac}       
\usepackage{microtype}      

\usepackage{graphicx}
\usepackage{amsmath}
\usepackage{amssymb}
\usepackage{subcaption}
\usepackage[capitalise]{cleveref}
\usepackage{rotating}

\usepackage{xcolor}

\usepackage{pifont}
\newcommand{\degree}{$^{\circ}$}

\usepackage{caption}
\usepackage{tabularx} 
\newcolumntype{Y}{>{\centering\arraybackslash}X}

\usepackage{booktabs}
\usepackage{multirow}

\usepackage{lipsum}

\title{Color Equivariant Convolutional Networks}

\author{
  Attila Lengyel \quad Ombretta Strafforello \quad Robert-Jan Bruintjes \\ \textbf{Alexander Gielisse} \quad  \textbf{Jan van Gemert}  \\
  Computer Vision Lab\\
  Delft University of Technology\\
  Delft, The Netherlands \\
}

\begin{document}

\maketitle

\begin{abstract}
  Color is a crucial visual cue readily exploited by Convolutional Neural Networks (CNNs) for object recognition. However, CNNs struggle if there is data imbalance between color variations introduced by accidental recording conditions. Color invariance addresses this issue but does so at the cost of removing all color information, which sacrifices discriminative power. In this paper, we propose Color Equivariant Convolutions (CEConvs), a novel deep learning building block that enables shape feature sharing across the color spectrum while retaining important color information. We extend the notion of equivariance from geometric to photometric transformations by incorporating parameter sharing over hue-shifts in a neural network. We demonstrate the benefits of CEConvs in terms of downstream performance to various tasks and improved robustness to color changes, including train-test distribution shifts. Our approach can be seamlessly integrated into existing architectures, such as ResNets, and offers a promising solution for addressing color-based domain shifts in CNNs.
\end{abstract}


\section{Introduction}
\label{sec:intro}

\begin{figure*}
\centering
\begin{subfigure}{0.53\textwidth}
\centering
\includegraphics[width=\linewidth]{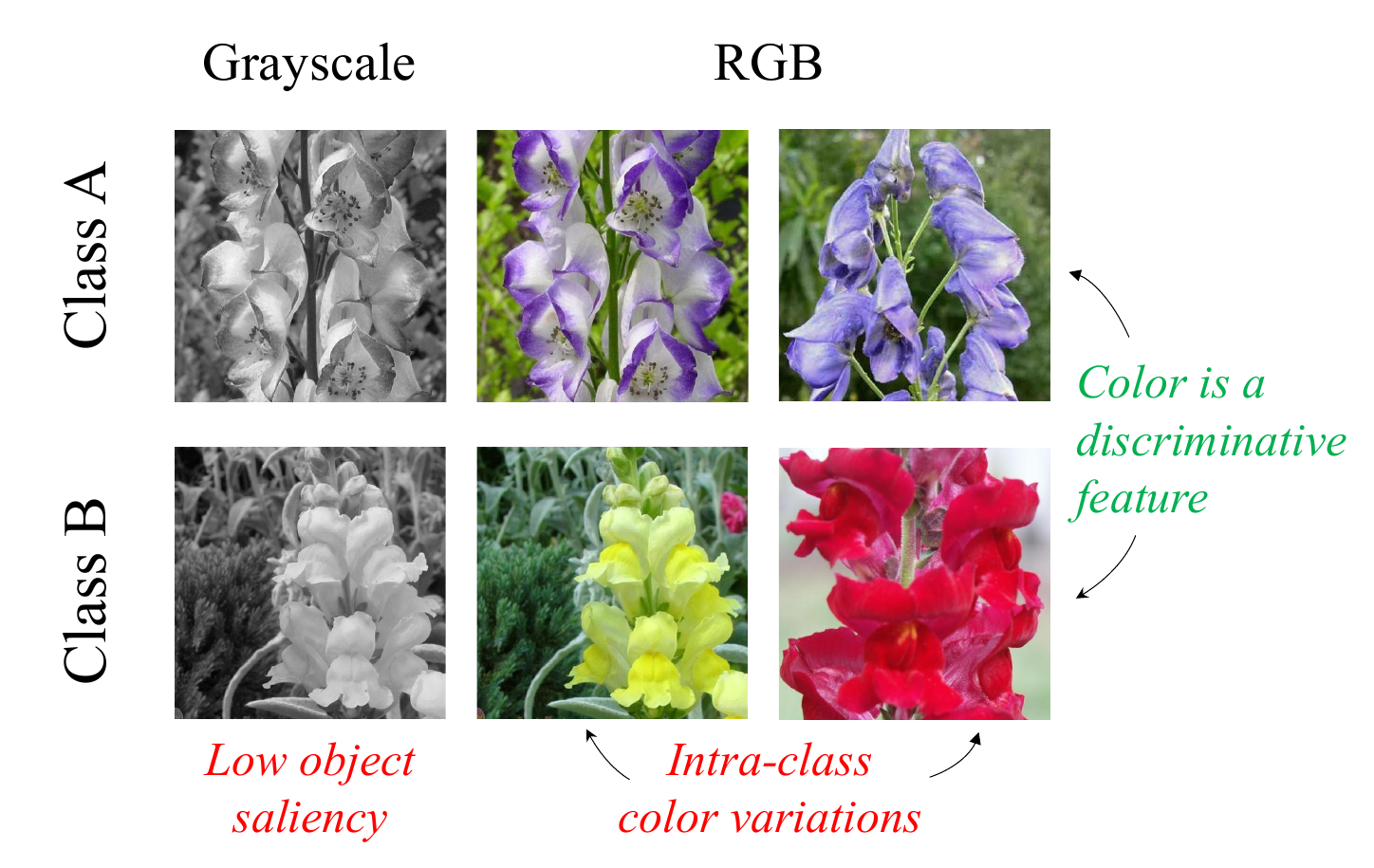}
\caption{}
\label{fig:figure-1}
\end{subfigure}%
\begin{subfigure}{0.47\textwidth}
\centering
\includegraphics[width=\linewidth]{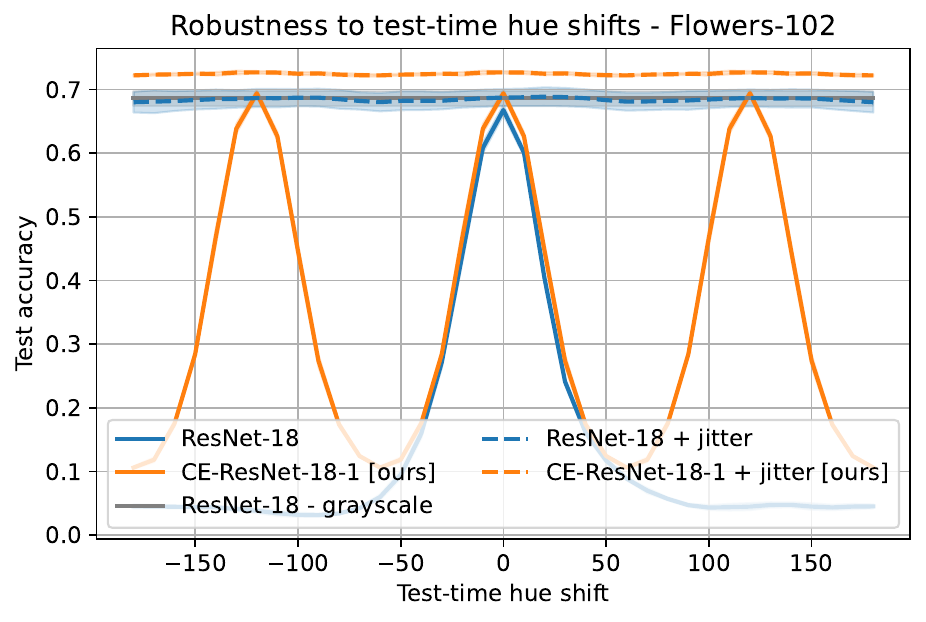}
\caption{}
\label{fig:imagenet_results}
\end{subfigure}
\caption{
Color plays a significant role in object recognition. (a) The absence of color makes flowers less distinct from their background and thus harder to classify. The characteristic purple-blue color of the Monkshood (Class A) enables a clear distinction from the Snapdragon (Class B) \cite{Nilsback08}. On the other hand, relying too much on colors might negatively impact recognition to color variations within the same flower class.
(b) Image classification performance on the Flower-102 dataset \cite{Nilsback08} under a gradual variation of the image hue. Test-time hue shifts degrade the performance of CNNs (ResNet-18) drastically. Grayscale images and color augmentations result in invariance to hue variations, but fail to capture all the characteristic color features of flowers.  Our color equivariant network (CE-ResNet-18-1) enables feature sharing across the color spectrum, which helps generalise to underrepresented colors in the dataset, while preserving discriminative color information, improving classification for unbalanced color variations. 
}
\end{figure*}

Color is a powerful cue for visual object recognition. Trichromatic color vision in primates may have developed to aid the detection of ripe fruits against a background of green foliage~\cite{osorio1996colour, regan2001fruits}. The benefit of color vision here is two-fold: not only does color information improve foreground-background segmentation by rendering foreground objects more salient, color also allows diagnostics, e.g. identifying the type (orange) and ripeness (green) where color is an intrinsic property facilitating recognition~\cite{bramo2012the}, as illustrated in \cref{fig:figure-1}. Convolutional neural networks (CNNs) too exploit color information by learning color selective features that respond differently based on the presence or absence of a particular color in the input~\cite{rafegas2018color}. 

Unwanted color variations,  however, can be introduced by accidental scene recording conditions such as illumination changes~\cite{lengyel2021zero, sakaridis2019guided}, or by low color-diagnostic objects occurring in a variety of colors, making color no longer a discriminative feature but rather an undesired source of variation in the data. Given a sufficiently large training set that encompasses all possible color variations, a CNN learns to become robust by learning color invariant and equivariant features from the available data~\cite{olah2020an, olah2020naturally}. However, due to the long tail of the real world it is almost impossible to collect balanced training data for all scenarios. This naturally leads to  color distribution shifts between training and test time, and an imbalance in the training data where less frequently occurring colors are underrepresented. As CNNs often fail to generalize to out-of-distribution test samples, this can have significant impact on many real-world applications, e.g. a model trained mostly on red cars may struggle to recognize the exact same car in blue.

\textit{Color invariance} addresses this issue through features that are by design invariant to color changes and therefore generalize better under appearance variations~\cite{finlayson2006on, geusebroek2001color}. However, color invariance comes at the loss of discriminative power as valuable color information is removed from the model's internal feature representation~\cite{gevers2012color}. We therefore propose to equip models with the less restrictive \textit{color equivariance} property, where features are explicitly shared across different colors through a hue transformation on the learned filters. This allows the model to generalize across different colors, while at the same time also retaining important color information in the feature representation.

An RGB pixel can be decomposed into an orthogonal representation by the well-known hue-saturation-value (HSV) model, where hue represents the chromaticity of a color. In this work we extend the notion of equivariance from geometric to photometric transformations by hard-wiring parameter sharing over hue-shifts in a neural network. More specifically, we build upon the seminal work of Group Equivariant Convolutions~\cite{cohen2016group} (GConvs), which implements equivariance to translations, flips and rotations of multiples of 90 degrees, and formulates equivariance using the mathematical framework of symmetry groups. We introduce Color Equivariant Convolutions (CEConvs) as a novel deep learning building block, which implements equivariance to the $H_n$ symmetry group of discrete hue rotations. CEConvs share parameters across hue-transformed filters in the input layer and store color information in hue-equivariant feature maps.

CEConv feature maps contain an additional dimension compared to regular CNNs, and as a result, require larger filters and thus more parameters for the same number of channels. To evaluate equivariant architectures, it is common practice to reduce the width of the network to match the parameter count of the baseline model. However, this approach introduces a trade-off between equivariance and model capacity, where particularly in deeper layers the quadratic increase in parameter count of CEConv layers makes equivariance computationally expensive. We therefore investigate hybrid architectures, where early color invariance is introduced by pooling over the color dimension of the feature maps. Note that early color invariance is maintained throughout the rest of the network, despite the use of regular convolutional layers after the pooling operation. Limiting color equivariant filters to the early layers is in line with the findings that early layers tend to benefit the most from equivariance~\cite{bruintjes2023affects} and learn more color selective filters~\cite{olah2020naturally, rafegas2018color}.

We rigorously validate the properties of CEConvs empirically through precisely  controlled synthetic experiments, and evaluate the performance of color invariant and equivariant ResNets on various more realistic classification benchmarks. Moreover, we investigate the combined effects of color equivariance and color augmentations. Our experiments show that CEConvs perform on par or better than regular convolutions, while at the same time significantly improving the robustness to test-time color shifts, and is complementary to color augmentations.

The main contributions of this paper can be summarized as follows:
\begin{itemize}
	\item We show that convolutional neural networks benefit from using color information, and at the same time are not robust to color-based domain shifts.
    \item We introduce Color Equivariant Convolutions (CEConvs), a novel deep learning building block that allows feature sharing between colors and can be readily integrated into existing architectures such as ResNets.
    \item We demonstrate that CEConvs improve robustness to train-test color shifts in the input.
\end{itemize}
All code and experiments are made publicly available on {\small\url{https://github.com/Attila94/CEConv}}.
\section{Related work}

\paragraph{Equivariant architectures}
Translation equivariance is a key property of convolutional neural networks (CNNs)~\cite{kayhan2020translation, lecun1998gradient}: shifting the input to a convolution layer results in an equally shifted output feature map. This allows CNNs to share filter parameters over spatial locations, which improves both parameter and data efficiency as the model can generalize to new locations not covered by the training set. A variety of methods have extended equivariance in CNNs to other geometric transformations~\cite{rath2020boosting}, including the seminal Group Equivariant Convolutions~\cite{cohen2016group} for rotations and flips, and other works concerning rotations~\cite{bekkers2018roto, lengyel2021exploiting, weiler2018learning}, scaling~\cite{sosnovik2020scale, worrall2019deep} and arbitrary Lie groups~\cite{macdonald2022enabling}. Yet to date, equivariance to photometric transformations has remained largely unexplored. Offset equivariant networks~\cite{cotogni2022offset} constrain the trainable parameters such that an additive bias to the RGB input channels results in	 an equal bias in the output logits. By applying a log transformation to the input the network becomes equivariant to global illumination changes according to the Von Kries model~\cite{finlayson1993diagonal}. In this work we explore an alternative approach to photometric equivariance inspired by the seminal Group Equivariant Convolution~\cite{cohen2016group} framework.

\paragraph{Color in CNNs}
Recent research has investigated the internal representation of color in Convolutional Neural Networks (CNNs), challenging the traditional view of CNNs as black boxes. For example, \cite{rafegas2017color, rafegas2018color} introduces the Neuron Feature visualization technique and characterizes neurons in trained CNNs based on their color selectivity, assessing whether a neuron activates in response to the presence of color in the input. The findings indicate that networks learn highly color-selective neurons across all layers, emphasizing the significance of color as a crucial visual cue. Additionally, \cite{rafegas2020understanding} classifies neurons based on their class selectivity and observes that early layers contain more class-agnostic neurons, while later layers exhibited high class selectivity. A similar study has been performed in~\cite{engilberge2017color}, further supporting these findings. \cite{olah2020an, olah2020naturally} investigate learned symmetries in an InceptionV1 model trained on ImageNet~\cite{deng2009imagenet} and discover filters that demonstrated equivariance to rotations, scale, hue shifts, and combinations thereof. These results motivate color equivariance as a prior for CNNs, especially in the first layers. Moreover, in this study, we will employ the metrics introduced by~\cite{rafegas2018color} to provide an explanation for several of our own findings.

\paragraph{Color priors in deep learning}
Color is an important visual discriminator~\cite{funt1995color,GeversSIST2012,van2011exploiting}. In classical computer vision, color invariants are used to extract features from an RGB image that are more consistent under illumination changes~\cite{finlayson2006on, geusebroek2001color, gevers2012color}. Recent studies have explored using color invariants as a preprocessing step to deep neural networks~\cite{alshammari2018on, maxwell2019real} or incorporating them directly into the architecture itself~\cite{lengyel2021zero}, leading to improved robustness against time-of-day domain shifts and other illumination-based variations in the input. Capsule networks~\cite{hinton2018matrix, sabour2017dynamic}, which use groups of neurons to represent object properties such as pose and appearance, have shown encouraging results in image colorization tasks~\cite{ozbulak2019image}. Quaternion networks~\cite{gaudet2018deep, zhu2018quaternion} represent RGB color values using quaternion notation, and employ quaternion convolutional layers resulting in moderate improvements in image classification and inpainting tasks. Building upon these advancements, we contribute to the ongoing research on integrating color priors within deep neural architectures.

\section{Color equivariant convolutions}

\subsection{Group Equivariant Convolutions}
A CNN layer $\Phi$ is equivariant to a symmetry group $G$ if for all transformations $g \in G$ on the input $x$ the resulting feature mapping $\Phi(x)$ transforms similarly, i.e., first doing a  transformation and then the mapping is similar to first doing the mapping and then the transformation. Formally, equivariance is defined as 
\begin{align}
\Phi(T_gx) = T'_g \Phi(x), \quad \forall g \in G,
\end{align}
where $T_g$ and $T'_g$ are the transformation operators of group action $g$ on the input and feature space, respectively. Note that $T_g$ and $T'_g$ can be identical, as is the case for translation equivariance where shifting the input results in an equally shifted feature map, but do not necessarily need to be. A special case of equivariance is invariance, where $T'_g$ is the identity mapping and the input transformation leaves the feature map unchanged:
\begin{align}
\Phi(T_gx) = \Phi(x),  \quad \forall g \in G.
\end{align}

We use the definition from~\cite{cohen2016group} to denote the $i$-th output channel of a standard convolutional layer $l$ in terms of the correlation operation $(\star)$ between a set of feature maps $f$ and $C^{l+1}$ filters $\psi$:
\begin{align}
    [f \star \psi^i](x) = \sum_{y \in \mathbb{Z}^2} \sum_{c=1}^{C^l} f_c(y)\psi_c^i(y-x).
\end{align}

Here $f : \mathbb{Z}^2 \rightarrow \mathbb{R}^{C^l}$ and $\psi^i : \mathbb{Z}^2 \rightarrow \mathbb{R}^{C^l}$ are functions that map pixel locations $x$ to a $C^l$-dimensional vector. This definition can be extended to groups by replacing the translation $x$ by a group action $g$:
\begin{align}
    [f \star \psi^i](g) &= \sum_{y \in \mathbb{Z}^2} \sum_{c}^{C^l} f_c(y)\psi^i_c(g^{-1}y)
\end{align}
As the resulting feature map $f \star \psi^i$ is a function on G rather than $\mathbb{Z}^2$, the inputs and filters of all hidden layers should also be defined on $G$:
\begin{align}
    [f \star \psi^i](g) &= \sum_{h \in G} \sum_{c}^{C^l} f_c(h)\psi^i_c(g^{-1}h)
\end{align}
Invariance to a subgroup can be achieved by applying a pooling operation over the corresponding cosets. For a more detailed introduction to group equivariant convolutions, please refer to~\cite{bronstein2021geometric, cohen2016group}.

\subsection{Color Equivariance}
We define color equivariance as equivariance to hue shifts. The HSV color space encodes hue by an angular scalar value, and a hue shift is performed as a simple additive offset followed by a modulo operator. When projecting the HSV representation into three-dimensional RGB space, the same hue shift becomes a rotation along the $[1,1,1]$ diagonal vector. 

We formulate hue equivariance in the framework of group theory by defining the group $H_n$ of multiples of $360/n$-degree rotations about the $[1,1,1]$ diagonal vector in $\mathbb{R}^3$ space. $H_n$ is a subgroup of the $SO(3)$ group of all rotations about the origin of three-dimensional Euclidean space. We can parameterize $H$ in terms of integers $k, n$ as
\begin{align}
\label{eq:rotation_matrix}
\begin{split}
H_n(k) = 
\begin{bmatrix}
\cos(\frac{2 k \pi}{n}) + a & a - b & a + b \\
a + b & \cos(\frac{2 k \pi}{n}) + a & a - b \\
a - b & a + b & \cos(\frac{2 k \pi}{n}) + a
\end{bmatrix}
\end{split}
\end{align}
with $n$ the total number of discrete rotations in the group, $k$ the rotation, $a = \frac{1}{3} - \frac{1}{3}\cos(\frac{2 k \pi}{n})$ and $b = \sqrt{\frac{1}{3}} * \sin(\frac{2 k \pi}{n})$. The group operation is matrix multiplication which acts on the continuous $\mathbb{R}^3$ space of RGB pixel values.
The derivation of $H_n$ is provided in \cref{app:derivation}.

\paragraph{Color Equivariant Convolution (CEConv)}
Let us define the group $G = \mathbb{Z}^2 \times H_n$, which is a direct product of the $\mathbb{Z}^2$ group of discrete 2D translations and the $H_n$ group of discrete hue shifts. 
We can then define the Color Equivariant Convolution (CEConv) in the input layer as:
\begin{align}
    [ f \star \psi^i ](x, k) &= \sum_{y \in \mathbb{Z}^2} \sum_{c=1}^{C^l} f_c(y) \cdot H_n(k)\psi^i_c(y-x).
\end{align}
We furthermore introduce the operator $\mathcal{L}_g = \mathcal{L}_{(t, m)}$ including translation $t$ and hue shift $m$ acting on input $f$ defined on the plane $\mathbb{Z}^2$:
\begin{align}
    [\mathcal{L}_{g} f](x) = [\mathcal{L}_{(t, m)} f](x) = H_n(m) f(x-t)
\end{align}
Since $H_n$ is an orthogonal matrix, the dot product between a hue shifted input $H_n f$ and a filter $\psi$ is equal to the dot product between the original input $f$ and the inverse hue shifted filter $H_n^{-1} \psi$:
\begin{equation}
\label{eq:hue_rotation_proof}
    H_n f \cdot \psi = (H_n f)^T \psi = f^T H_n^T \psi = f \cdot H_n^T \psi = f \cdot H_n^{-1} \psi.
\end{equation}
Then the equivariance of the CEConv layer can be derived as follows (using $C^l=1$ for brevity):
\begin{align}
\begin{split}
    [ [\mathcal{L}_{(t, m)} f] \star \psi^i ](x, k) &= \sum_{y \in \mathbb{Z}^2} H_n(m) f(y-t) \cdot H_n (k)\psi^i(y-x) \\
    &= \sum_{y \in \mathbb{Z}^2} f(y) \cdot H_n(m)^{-1} H_n(k) \psi^i(y - (x - t)) \\
    &= \sum_{y \in \mathbb{Z}^2} f(y) \cdot H_n(k-m) \psi^i(y - (x - t)) \\
    &= [f \star \psi^i](x-t, k-m) \\
    &= [\mathcal{L'}_{(t, m)} [f \star \psi^i]](x, k)
\end{split}
\end{align}
Since input $f$ and feature map $[f \star \psi]$ are functions on $\mathbb{Z}^2$ and $G$, respectively, $\mathcal{L}_{(t, k)}$ and $\mathcal{L'}_{(t, k)}$ represent two equivalent operators acting on their respective groups. For all subsequent hidden layers the input $f$ and filters $\psi^i$ are functions on $G$ parameterized by $x,k$, and the hidden layer for CEConv is defined as:
\begin{align}
    [ f \star \psi^i ](x,k) &= \sum_{y \in \mathbb{Z}^2} \sum_{r=1}^n \sum_{c=1}^{C^l} f_c(y,r) \cdot \psi^i_c(y-x, (r-k) {\scriptstyle\%} n),
\end{align}
where $n$ is the number of discrete rotations in the group and ${\scriptstyle\%}$ is the modulo operator.
In practice, applying a rotation to RGB pixels will cause some pixel values to fall outside of the RGB cube, which will then have to be reprojected within the cube. Due to this discrepancy, \cref{eq:hue_rotation_proof} only holds approximately, though in practice this has only limited consequences, as we empirical show in \cref{app:ablation}.

\subsection{Implementation}
\paragraph{Tensor operations} We implement CEConv similarly to GConv~\cite{cohen2016group}. GConv represents the pose associated with the added spatial rotation group by extending the feature map tensor $X$ with an extra dimension $G^l$ to size $[C^l, G^l, H, W]$, denoting the number of channels, transformations that leave the origin invariant, and height and width of the feature map at layer $l$, respectively (batch dimension omitted). Similarly, a GConv filter $\Tilde{F}$ with spatial extent $k$ is of size $[C^{l+1}, G^{l+1}, C^l, G^l, k, k]$. The GConv is then defined in terms of tensor multiplication operations as:
\begin{align}
\label{eq:gconv_tensor}
    X_{c',g',:,:}^{l+1} &= \sum_c^{C^{l}} \sum_g^{G^l} \Tilde{F}^l_{c',g',c,g,:,:} \star X_{c,g,:,:}^l,
\end{align}
where $(:)$ denotes tensor slices. Note that in the implementation, a GConv filter $F$ only contains $[C^{l+1}, C^l, G^l, k, k]$ unique parameters - the extra $G^{l+1}$ dimension is made up of transformed copies of $F$.

As the RGB input to the network is defined on $\mathbb{Z}^2$, we have $G^1=1$ and $\Tilde{F}$ has size $[C^{l+1}, G^{l+1}, 3, 1, k, k]$. The transformed copies in $G^{l+1}$ are computed by applying the rotation matrix from \cref{eq:rotation_matrix}:
\begin{align}
\label{eq:ceconv_inputfilter}
    \Tilde{F}^1_{c',g',:,1,u,v} &= H_n(g') F^1_{c',:,1,u,v}.
\end{align}
In the hidden layers $\Tilde{F}$ contains cyclically permuted copies of $F$:
\begin{align}
\label{eq:ceconv_hiddenfilter}
    \Tilde{F}^l_{c',g',c,g,u,v} &= F^l_{c',c,(g+g'){\scriptstyle\%}n,u,v}.
\end{align}
Furthermore, to explicitly share the channel-wise spatial kernel over $G^l$~\cite{lengyel2021exploiting}, filter $F$ is decomposed into a spatial component $S$ and a pointwise component $P$ as follows:
\begin{align}
\label{eq:decomposition}
    F^l_{c',c,g,u,v} &= S_{c',c,1,u,v} \cdot P_{c',g',c,g,1,1}
\end{align}
$F$ is precomputed in each forward step prior to the convolution operation in \cref{eq:gconv_tensor}.

\paragraph{Input normalization} is performed using a single value for the mean and standard deviations rather than per channel, as is commonly done for standard CNNs. Channel-wise means and standard deviations break the equivariance property of CECNN as a hue shift could no longer be defined as a rotation around the $[1,1,1]$ diagonal. Experiments have shown that using a single value for all channels instead of channel-wise normalization has no effect on the performance.

\paragraph{Compute efficiency}
CEConvs create a factor $|H_n|$ more feature maps in each layer. Due to the decomposition in \cref{eq:decomposition}, the number of multiply-accumulate (MAC) operations increase by only a factor $\frac{|H_n|^2}{k^2} + |H_n|$, and the number of parameters by a factor $\frac{|H_n|}{k^2}+1$. See \cref{app:configs} for an overview of parameter counts and MAC operations.



\section{Experiments}

\subsection{When is color equivariance useful?}
Color equivariant convolutions share shape information across different colors while preserving color information in the group dimension. To demonstrate when this property is useful we perform two controlled toy experiments on variations of the MNIST~\cite{deng2012mnist} dataset. We use the Z2CNN architecture from~\cite{cohen2016group}, and create a color equivariant version of the network called CECNN by replacing all convolutional layers by CEConvs with three rotations of 120\degree. The number of channels in CECNN is scaled such as to keep the number of parameters approximately equal to the Z2CNN. We also create a color invariant CECNN by applying coset max-pooling after the final CEConv layer, and a color invariant Z2CNN by converting the inputs to grayscale. All experiments are performed using the Adam~\cite{kingma2014adam} optimizer with a learning rate of 0.001 and the OneCycle learning rate scheduler. No data augmentations are used. We report the average performance over ten runs with different random initializations.

\begin{figure*}
\centering
\begin{subfigure}{0.5833\textwidth}
\centering
\includegraphics[width=\linewidth]{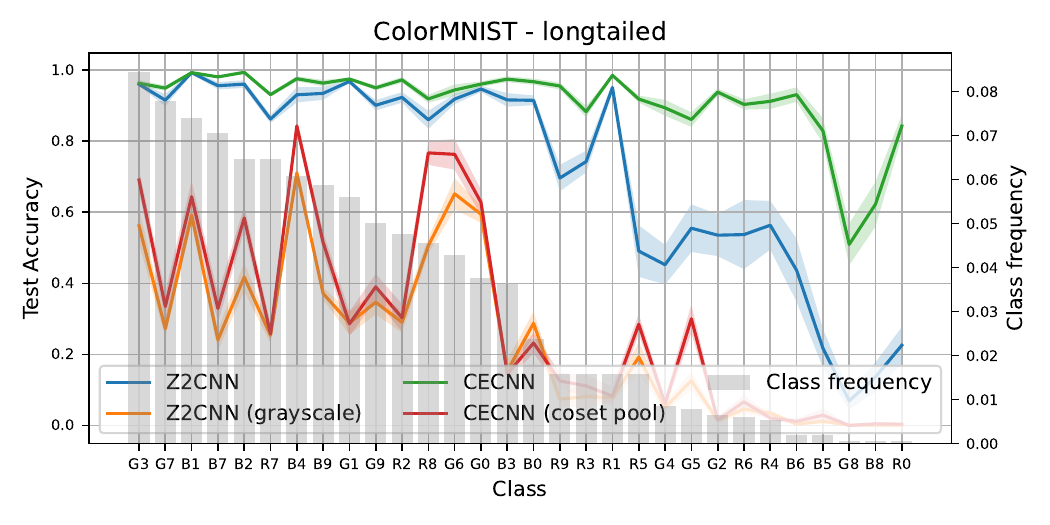}
\caption{}
\label{fig:colormnist_longtailed}
\end{subfigure}%
\begin{subfigure}{0.4166\textwidth}
\centering
\includegraphics[width=\linewidth]{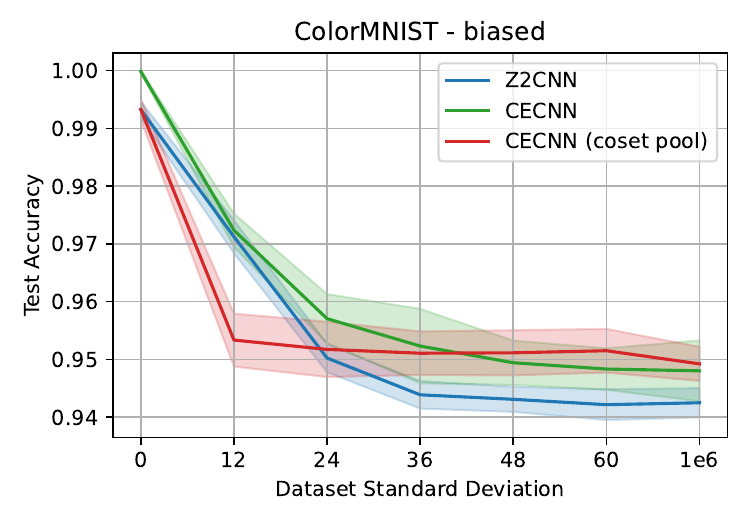}
\caption{}
\label{fig:colormnist_biased}
\end{subfigure}
\caption{Color equivariant convolutions efficiently share shape information across different colors. CECNN outperforms a vanilla network in both a long-tailed class imbalance setting (a), where MNIST digits are to be classified based on both shape and color, and a color biased setting (b), where the color of each class $c$ is sampled according to $\theta_d \sim \mathcal{N}(\theta_c, \sigma)$.}
\end{figure*}

\paragraph{Color imbalance} is simulated by \textit{long-tailed ColorMNIST}, a 30-class classification problem where digits occur in three colors on a gray background, and need to be classified by both number (0-9) and color (red, green, blue). The number of samples per class is drawn from a power law distribution resulting in a long-tailed class imbalance. Sharing shape information across colors is beneficial as a certain digit may occur more frequently in one color than in another. The train set contains a total of 1,514 training samples and the test set is uniformly distributed with 250 samples per class. The training set is visualized in \cref{app:visualizations}. We train all four architectures on the dataset for 1000 epochs using the standard cross-entropy loss. The train set distribution and per-class test accuracies for all models are shown in \cref{fig:colormnist_longtailed}. With an average accuracy of $91.35 \pm 0.40 \%$ the CECNN performs significantly better than the CNN with $71.59 \pm 0.61\%$. The performance increase is most significant for the classes with a low sample size, indicating that CEConvs are indeed more efficient in sharing shape information across different colors. The color invariant Z2CNN and CECNN networks, with an average accuracy of $24.19 \pm 0.53 \%$ and $29.43 \pm 0.46 \%$, respectively, are unable to discriminate between colors. CECNN with coset pooling is better able to discriminate between foreground and background and therefore performs slightly better. We repeated the experiment with a weighted loss and observed no significantly different results. We have also experimented with adding color jitter augmentations, which makes solving the classification problem prohibitive, as color is required. See \cref{app:colormnist_additional} for both detailed results on both experiments.

\paragraph{Color variations} are simulated by \textit{biased ColorMNIST}, a 10-class classification problem where each class $c$ has its own characteristic hue $\theta_c$ defined in degrees, distributed uniformly on the hue circle. The exact color of each digit $x$ is sampled according to $\theta_x \sim \mathcal{N}(\theta_c, \sigma)$. We generate multiple datasets by varying $\sigma$ between 0 and $10^6$, where $\sigma = 0$ results in a completely deterministic color for each class and $\sigma = 10^6$ in an approximately uniform distribution for $\theta_x$. For small $\sigma$, color is thus highly informative of the class, whereas for large $\sigma$ the classification needs to be performed based on shape. The dataset is visualized in \cref{app:visualizations}. We train all models on the train set of 1.000 samples for 1500 epochs and evaluate on the test set of 10.000 samples. The test accuracies for different $\sigma$ are shown in \cref{fig:colormnist_biased}. CECNN outperforms Z2CNN across all standard deviations, indicating CEConvs allow for a more efficient internal color representation. The color invariant CECNN network outperforms the equivariant CECNN model from $\sigma \geq 48$. Above this value color is no longer informative for the classification task and merely acts as noise unnecessarily consuming model capacity, which is effectively filtered out by the color invariant networks. The results of the grayscale Z2CNN are omitted as they are significantly worse, ranging between $89.89 \%$ ($\sigma = 0$) and $79.94$ ($\sigma = 10^6$). Interestingly, CECNN with coset pooling outperforms the grayscale Z2CNN. This is due to the fact that a CECNN with coset pooling is still able to distinguish between small color changes and therefore can partially exploit color information. Networks trained with color jitter are unable to exploit color information for low $\sigma$; see \cref{app:colormnist_additional} for detailed results.

\subsection{Image classification}
\label{sec:image_classification}

\paragraph{Setup} We evaluate our method for robustness to color variations on several natural image classification datasets, including CIFAR-10 and CIFAR-100~\cite{cifar10}, Flowers-102~\cite{Nilsback08}, STL-10~\cite{coates2011analysis}, Oxford-IIIT Pet~\cite{parkhi12a}, Caltech-101~\cite{li_andreeto_ranzato_perona_2022}, Stanford Cars~\cite{krause2013object} and ImageNet~\cite{deng2009imagenet}. We train a baseline and color equivariant (CE-)ResNet~\cite{he2016residual} with 3 rotations and evaluate on a range of test sets where we gradually apply a hue shift between -180\degree~and 180\degree. For high-resolution datasets (all except CIFAR) we train a ResNet-18 architecture and use default ImageNet data augmentations: we scale to 256 pixels, random crop to 224 pixels and apply random horizontal flips. For the CIFAR datasets we use the ResNet-44 architecture and augmentations from~\cite{cohen2016group}, including random horizontal flips and translations of up to 4 pixels. We train models both with and without color jitter augmentation to separately evaluate the effect of equivariance and augmentation. The CE-ResNets are downscaled in width to match the parameter count of the baseline ResNets. We have also included AugMix~\cite{hendrycks2020augmix} and CIConv~\cite{lengyel2021zero} as baselines for comparison. Training is performed for 200 epochs using the Adam~\cite{Kingma2014AdamAM} optimizer with a learning rate of 0.001 and the OneCycle learning rate scheduler. All our experiments use PyTorch and run on a single NVIDIA A40 GPU.

\paragraph{Hybrid networks}
In our toy experiments we enforce color equivariance throughout the network. For real world datasets however, we anticipate that the later layers of a CNN may not benefit from enforcing parameter sharing between colors, if the classes of the dataset are determined by color specific features. We therefore evaluate hybrid versions of our color equivariance networks, denoted by an integer suffix for the number of ResNet stages, out of a possible four, that use CEConvs.

\begin{table}[ht]
\setlength{\tabcolsep}{2pt}
\small
    \centering
    \begin{tabularx}{\linewidth}{@{}lYYYYYYYY@{}}
        \toprule
        \textit{Original test set} & \textbf{Caltech} & \textbf{C-10} & \textbf{C-100} & \textbf{Flowers} & \textbf{Ox-Pet} & \textbf{Cars} & \textbf{STL10} & \textbf{ImageNet} \\ \midrule
        Baseline            & 71.61 & 93.69 & 71.28 & 66.79 & 69.87 & 76.54 & 83.80 & 69.71 \\
        CIConv-W            & \textbf{72.85} & 75.26 & 38.81 & \textbf{68.71} & 61.53 & \textbf{79.52} & 80.71 & 65.81 \\
        CEConv              & 70.16 & 93.71 & 71.37 & 68.18 & 70.24 & 76.22 & 84.24 & 66.85 \\
        CEConv-2            & 71.50 & \textbf{93.94} & \textbf{72.20} & 68.38 & \textbf{70.34} & 77.06 & \textbf{84.50} & \textbf{70.02} \\ \midrule
        Baseline + jitter   & 73.93 & 93.03 & 69.23 & 68.75 & 72.71 & 80.59 & 83.91 & 69.37 \\
        CIConv-W + jitter   & \textbf{74.38} & 77.49 & 42.27 & \textbf{75.05} & 64.23 & \textbf{81.56} & 81.88 & 65.95 \\
        CEConv + jitter     & 73.58 & 93.51 & 71.12 & 74.17 & \textbf{73.29} & 79.79 & 84.16 & 65.57 \\
        CEConv-2 + jitter   & 72.61 & \textbf{93.86} & \textbf{71.35} & 71.72 & 72.80 & 80.32 & \textbf{84.46} & \textbf{69.42} \\ \midrule
        Baseline + AugMix   & \textbf{71.92} & 94.13 & \textbf{72.64} & 75.49 & \textbf{76.02} & \textbf{82.32} & 84.99 & - \\
        CEConv + AugMix     & 70.74 & \textbf{94.22} & 72.48 & \textbf{78.10} & 75.90 & 80.81 & \textbf{85.46} & - \\ \midrule
        \multicolumn{5}{@{}l}{\textit{Hue-shifted test set}} \\ \midrule
        Baseline            & 51.14 & 85.26 & 47.01 & 13.41 & 37.56 & 55.59 & 67.60 & 54.72 \\
        CIConv-W            & \textbf{71.92} & 74.88 & 37.09 & \textbf{59.03} & \textbf{60.54} & \textbf{78.71} & \textbf{79.92} & \textbf{64.62} \\
        CEConv              & 62.17 & 90.90 & 59.04 & 33.33 & 54.02 & 67.16 & 78.25 & 56.90 \\
        CEConv-2            & 64.51 & \textbf{91.43} & \textbf{62.11} & 33.32 & 51.14 & 68.17 & 77.80 & 62.26 \\ \midrule
        Baseline + jitter   & 73.61 & 92.91 & 69.12 & 68.44 & 72.30 & 80.65 & 83.71 & 67.10 \\
        CIConv-W + jitter   & \textbf{74.40} & 77.28 & 42.30 & \textbf{75.66} & 63.93 & \textbf{81.44} & 81.54 & 65.03 \\
        CEConv + jitter     & 73.57 & 93.39 & 71.06 & 73.86 & \textbf{72.94} & 79.79 & 84.02 & 64.52 \\
        CEConv-2 + jitter   & 73.03 & \textbf{93.80} & \textbf{71.33} & 71.44 & 72.58 & 80.28 & \textbf{84.31} & \textbf{68.74} \\ \midrule
        Baseline + AugMix   & 51.82 & 88.03 & 51.39 & 15.99 & 48.04 & 68.69 & 72.19 & - \\
        CEConv + AugMix     & \textbf{62.29} & \textbf{91.68} & \textbf{60.75} & \textbf{41.43} & \textbf{62.27} & \textbf{73.59} & \textbf{80.17} & - \\ \bottomrule
    \end{tabularx}
    \caption{Classification accuracy in \% of vanilla vs. color equivariant (CE-)ResNets, evaluated both on the original and hue-shifted test sets. Color equivariant CNNs perform on par with vanilla CNNs on the original test sets, but are significantly more robust to test-time hue shifts.}
    \label{tab:classification}
    \vspace{-4mm}
\end{table}

\paragraph{Results} We report both the performance on the original test set, as well as the average accuracy over all hue shifts in \cref{tab:classification}. For brevity we only show the fully equivariant and hybrid-2 networks, a complete overview of the performances of all hybrid network configurations and error standard deviations can be found in \cref{app:cls}. Between the full color equivariant and hybrid versions of our CE-ResNets, at least one variant outperforms vanilla ResNets on most datasets on the original test set. On most datasets the one- or two-stage hybrid versions are the optimal CE-ResNets, providing a good trade-off between color equivariance and leaving the network free to learn color specific features in later layers. CE-ResNets are also significantly more robust to test-time hue shifts, especially when trained without color jitter augmentation. Training the CE-ResNets with color jitter further improves robustness, indicating that train-time augmentations complement the already hard-coded inductive biases in the network. We show the detailed performance on Flowers-102 for all test-time hue shifts in \cref{fig:imagenet_results}. The accuracy of the vanilla CNN quickly drops as a hue shift is applied, whereas the CE-CNN performance peaks at -120\degree, 0\degree and 120\degree. Applying train-time color jitter improves the CNN's robustness to the level of a CNN with grayscale inputs. The CE-CNN with color jitter outperforms all models for all hue shifts. Plots for other datasets are provided in \cref{app:test_hue_shift}.

\begin{figure*}[t]
\centering
\includegraphics[width=0.8\linewidth]{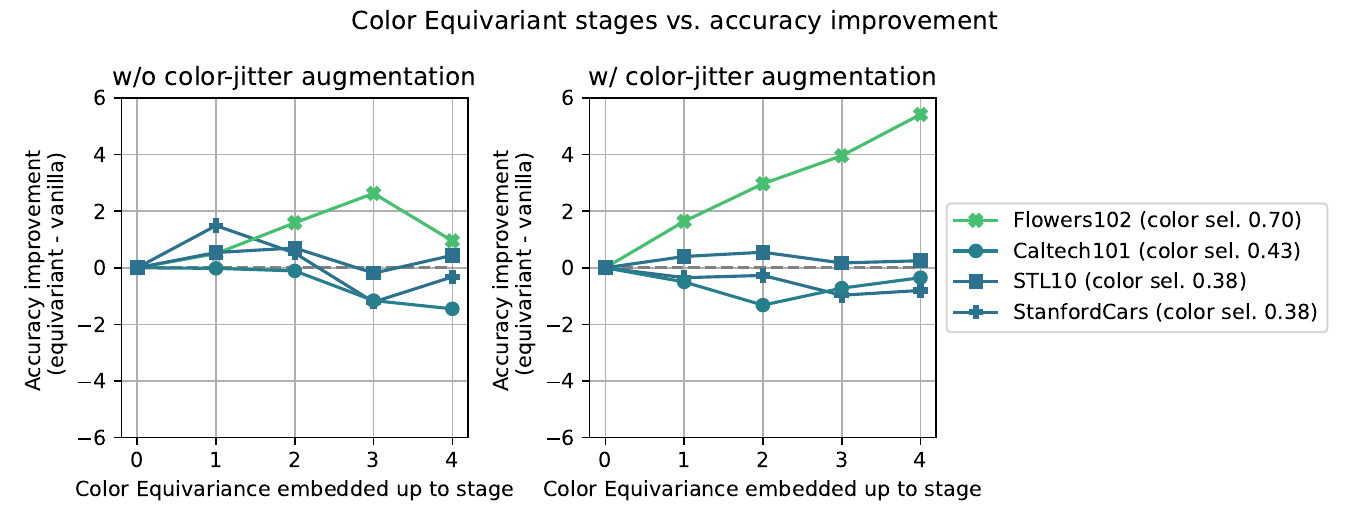}
\caption{Color selective datasets benefit from using color equivariance up to late stages, whereas less color selective datasets do not. We compute average color selectivity~\cite{rafegas2018color} of all neurons in the baseline CNN trained on each dataset, and plot the accuracy improvement of using color equivariance in hybrid and full models, coloring each graphed dataset for color selectivity.}
\label{fig:ColorSelectivity}
\vspace{-3mm}
\end{figure*}

\paragraph{Color selectivity}
To explore what affects the success of color equivariance, we investigate the \textit{color selectivity} of a subset of the studied datasets. We use the color selectivity measure from~\cite{rafegas2018color} and average across all neurons in the baseline model trained on each dataset. \cref{fig:ColorSelectivity} shows that color selective datasets benefit from using color equivariance up to late stages, whereas
less color selective datasets do not.



\begin{figure}[t]
    \centering
    \includegraphics[width=\linewidth]{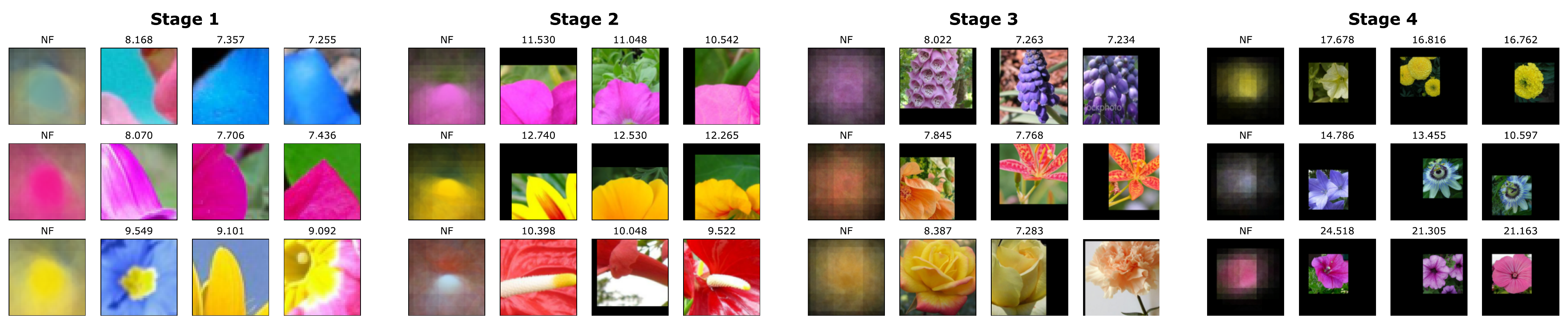}
    \caption{Neuron Feature~\cite{rafegas2018color} (NF) visualization with top-3 patches at different stages of a CE-ResNet18 trained on Flowers-102. Rows represent different rotations of the same filter. As expected, each row of a NF activates on the same shape in a different color.}
    \label{fig:neuronfeature}
\end{figure}

\paragraph{Feature representations of color equivariant CNNs} We use the Neuron Feature~\cite{rafegas2018color} (NF) visualization method to investigate the internal feature representation of the CE-ResNet. NF computes a weighted average of the $N$ highest activation input patches for each filter at a certain layer, as such representing the input patch that a specific neuron fires on. \cref{fig:neuronfeature} shows the NF ($N=50$) and top-3 input patches for filters at the final layers of stages 1-4 of a CE-ResNet18 trained on Flowers-102. Different rows represent different rotations of the same filter. As expected, each row of a NF activates on the same shape in a different color, demonstrating the color sharing capabilities of CEConvs. More detailed NF visualization are provided in \cref{app:nf_vis}.

\paragraph{Ablation studies}
We perform ablations to investigate the effect of the number of rotations, the use of group coset pooling, and the strength of train-time color jitter augmentations. In short, we find that a) increasing the number of hue rotations increases robustness to test-time hue shifts at the cost of a slight reduction in network capacity, b) removing group coset pooling breaks hue invariance, and c) hue equivariant networks require lower intensity color jitter augmentations to achieve the same test-time hue shift robustness and accuracy. The full results can be found in \cref{app:ablation}.
\section{Conclusion}

In this work, we propose Color Equivariant Convolutions (CEConvs) which enable feature sharing across colors in the data, while retaining discriminative power. Our toy experiments demonstrate benefits for datasets where the color distribution is long-tailed or biased. Our proposed fully equivariant CECNNs improve performance on datasets where features are color selective, while hybrid versions that selectively apply CEConvs only in early stages of a CNN benefit various classification tasks.

\paragraph{Limitations}
CEConvs are computationally more expensive than regular convolutions. For fair comparison, we have equalized the parameter cost of all models compared, at the cost of reducing the number of channels of CECNNs. In cases where color equivariance is not a useful prior, the reduced capacity hurts model performance, as reflected in our experimental results.

Pixel values near the borders of the RGB cube can fall outside the cube after rotation, and subsequently need to be reprojected. Due to this clipping effect the hue equivariance in \cref{eq:hue_rotation_proof} only holds approximately. As demonstrated empirically, this has only limited practical consequences, yet future work should investigate how this shortcoming could be mitigated.

\paragraph{Local vs. global equivariance}
The proposed CEConv implements local hue equivariance, i.e. it allows to model local color changes in different regions of an image separately. In contrast, global equivariance, e.g. by performing hue shifts on the full input image, then processing all inputs with the same CNN and combining representations at the final layer to get a hue-equivariant representation, encodes global equivariance to the entire image. While we have also considered such setup, initial experiments did not yield promising results. The theoretical benefit of local over global hue equivariance is that multiple objects in one image can be recognized equivariantly in any combination of hues - empirically this indeed proves to be a useful property.

\paragraph{Future work}
The group of hue shifts is but one of many possible transformations groups on images. CNNs naturally learn features that vary in both photometric and geometric transformations \cite{bruintjes2023affects, olah2020naturally}. Future work could combine hue shifts with geometric transformations such as roto-translation \cite{cohen2016group} and scaling \cite{sosnovik2021disco}. Also, other photometric properties could be explored in an equivariance setting, such as saturation and brightness.

Our proposed method rotates the hue of the inputs by a predetermined angle as encoded in a rotation matrix. Making this rotation matrix learnable could yield an inexact but more flexible type of color equivariance, in line with recent works on learnable equivariance~\cite{moskalev2022liegg,romero2021learning}. An additional line of interesting future work is to incorporate more fine-grained equivariance to continuous hue shifts, which is currently intractable within the GConv-inspired framework as the number multiply-accumulate operations grow quadratically with the number of hue rotations. 

\paragraph{Broader impact}
Improving performance on tasks where color is a discriminative feature could affect humans that are the target of discrimination based on the color of their skin. CEConvs ideally benefit datasets with long-tailed color distributions by increasing robustness to color changes, in theory reducing a CNN's reliance on skin tone as a discriminating factor. However, careful and rigorous evaluation is needed before such properties can be attributed to CECNNs with certainty.

\section*{Acknowledgements} This project is supported in part by NWO (project VI.Vidi.192.100).

{\small
\bibliographystyle{ieee_fullname}
\bibliography{egbib}
}


\newpage
\appendix
\appendix


\section{Derivation of $H_n$}
\label{app:derivation}

Rotation around an arbitrary unit vector $\mathbf{u}$ by angle $\theta$ can be decomposed into five simple steps~\cite{Cole2015}:
\begin{enumerate}
    \item rotating the vector such that it lies in one of the coordinate planes, e.g. $xz$ using $M_{xz}$;
    \item rotating the vector such that it lies on one of the coordinate axes, e.g. $x$ using $M_x$;
    \item rotating the point around vector $\mathbf{u}$ on axis $x$ using $R_x$;
    \item reversing the rotation in step 2. using $M^{-1}_x=M^T_x$;
    \item reversing the rotation in step 1. using $M^{-1}_{xz}=M^T_{xz}$.
\end{enumerate}
These operations can be combined into a single matrix:
\begin{align}
    R_{\mathbf{u},\theta} &= M^T_{xz}(M^T_x(R_{x,\theta}(M_{xz}(M_{xz})))) \\
    &= M^T_{xz}M^T_xR_{x,\theta}M_{xz}M_{xz} \\
    &= 
    \begin{bmatrix} 
    \cos \theta +u_x^2 \left(1-\cos \theta\right) &
    u_x u_y \left(1-\cos \theta\right) - u_z \sin \theta &
    u_x u_z \left(1-\cos \theta\right) + u_y \sin \theta \\
    u_y u_x \left(1-\cos \theta\right) + u_z \sin \theta &
    \cos \theta + u_y^2\left(1-\cos \theta\right) &
    u_y u_z \left(1-\cos \theta\right) - u_x \sin \theta \\
    u_z u_x \left(1-\cos \theta\right) - u_y \sin \theta &
    u_z u_y \left(1-\cos \theta\right) + u_x \sin \theta &
    \cos \theta + u_z^2\left(1-\cos \theta\right)
\end{bmatrix}.
\end{align}
Substituting $\mathbf{u}=[\frac{1}{\sqrt{3}},\frac{1}{\sqrt{3}},\frac{1}{\sqrt{3}}]$ yields
\begin{align}
    R_{\mathbf{u},\theta} &=
    \begin{bmatrix} 
    \cos \theta + \frac{1}{3} \left(1-\cos \theta\right) &
    \frac{1}{3} \left(1-\cos \theta\right) - \frac{1}{\sqrt{3}} \sin \theta &
    \frac{1}{3} \left(1-\cos \theta\right) + \frac{1}{\sqrt{3}} \sin \theta \\
    \frac{1}{3} \left(1-\cos \theta\right) + \frac{1}{\sqrt{3}} \sin \theta &
    \cos \theta + \frac{1}{3}\left(1-\cos \theta\right) &
    \frac{1}{3} \left(1-\cos \theta\right) - \frac{1}{\sqrt{3}} \sin \theta \\
    \frac{1}{3} \left(1-\cos \theta\right) - \frac{1}{\sqrt{3}} \sin \theta &
    \frac{1}{3} \left(1-\cos \theta\right) + \frac{1}{\sqrt{3}} \sin \theta &
    \cos \theta + \frac{1}{3}\left(1-\cos \theta\right)
\end{bmatrix}, 
\end{align}
and lastly, rearranging and substituting $\theta=\frac{2 k \pi}{n}$ results in
\begin{align}
\begin{split}
H_n(k) = 
\begin{bmatrix}
\cos(\frac{2 k \pi}{n}) + a & a - b & a + b \\
a + b & \cos(\frac{2 k \pi}{n}) + a & a - b \\
a - b & a + b & \cos(\frac{2 k \pi}{n}) + a
\end{bmatrix}.
\end{split}
\end{align}
with $n$ the total number of discrete rotations in the group, $k$ the rotation, $a = \frac{1}{3} - \frac{1}{3}\cos(\frac{2 k \pi}{n})$ and $b = \sqrt{\frac{1}{3}} * \sin(\frac{2 k \pi}{n})$.

\newpage

\section{ColorMNIST}
\subsection{Dataset visualization}
\label{app:visualizations}
\paragraph{Long-tailed ColorMNIST dataset}

\begin{figure}[ht!]
    \centering
    \includegraphics[width=\linewidth]{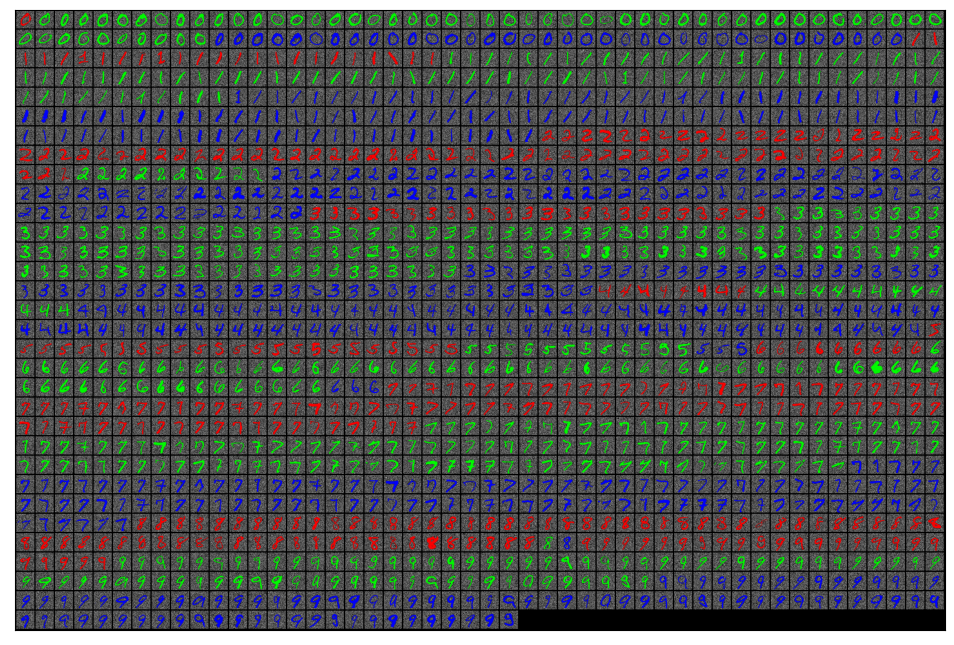}
    \caption{Long-tailed ColorMNIST. Note the strong class imbalance in the dataset. Best viewed in color.}
    \label{fig:colormnist_longtailed_samples}
\end{figure}

The training samples of the \textit{Longtailed ColorMNIST} dataset are depicted in \cref{fig:colormnist_longtailed_samples}, clearly indicating a class imbalance.

\begin{figure*}[ht!]
\centering
\begin{subfigure}{0.3\textwidth}
\centering
\includegraphics[width=\linewidth]{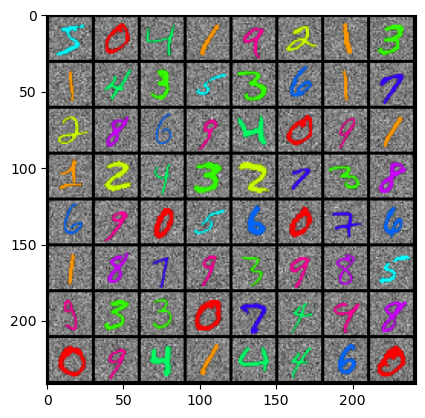}
\caption{}
\label{fig:colormnist_biased_samples_0}
\end{subfigure}%
\begin{subfigure}{0.3\textwidth}
\centering
\includegraphics[width=\linewidth]{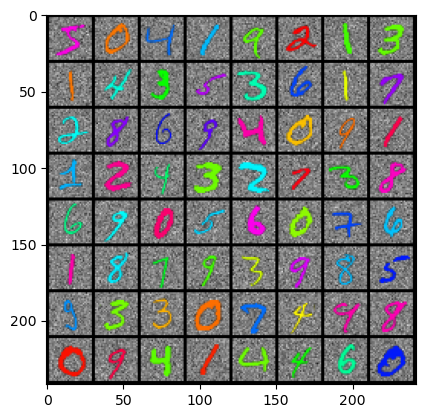}
\caption{}
\label{fig:colormnist_biased_samples_36}
\end{subfigure}
\caption{Samples from Biased ColorMNIST for $\sigma=0$ (a) and $\sigma=36$ (b), respectively. Best viewed in color.}
\label{fig:colormnist_biased_samples}
\end{figure*}

\paragraph{Biased ColorMNIST dataset}
A small subset of the samples of Biased ColorMNIST is shown in \cref{fig:colormnist_biased_samples} for $\sigma=0$ (a) and $\sigma=36$ (b), respectively. Note that the samples in (a) have a deterministic color, whereas in (b) exhibit some variation in hue.

\subsection{Additional experiments}
\label{app:colormnist_additional}
\begin{figure*}[ht!]
\centering
\begin{subfigure}{0.5833\textwidth}
\centering
\includegraphics[width=\linewidth]{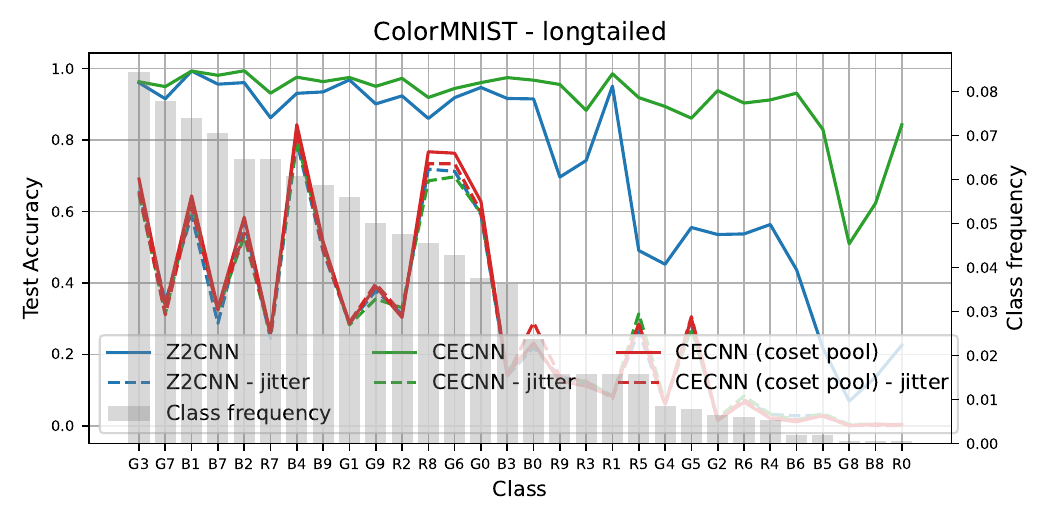}
\caption{}
\end{subfigure}%
\begin{subfigure}{0.4166\textwidth}
\centering
\includegraphics[width=\linewidth]{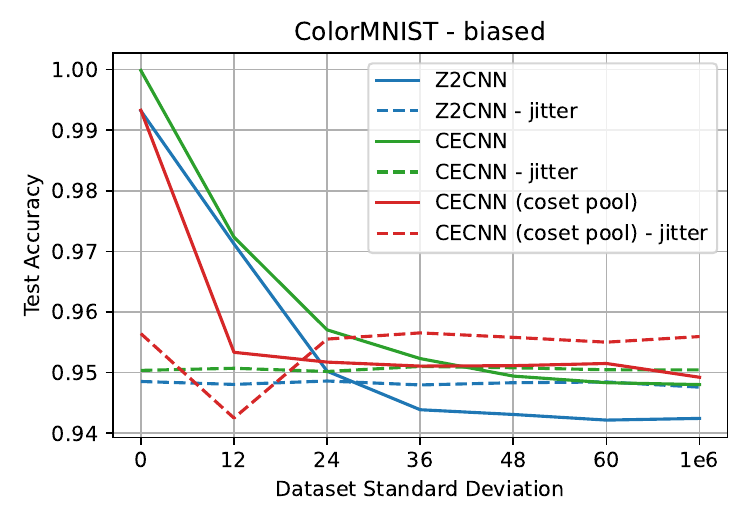}
\caption{}
\end{subfigure}
\caption{Color equivariant convolutions efficiently share shape information across different colors. CECNN outperforms a vanilla network in both a long-tailed class imbalance setting (a), where MNIST digits are to be classified based on both shape and color, and a color biased setting (b), where the color of each class $c$ is sampled according to $\theta_d \sim \mathcal{N}(\theta_c, \sigma)$.}
\label{fig:colormnist_jitter}
\end{figure*}

\paragraph{Results with color jitter augmentation}
We performed both ColorMNIST experiments with color jitter augmentations. The results are shown in \cref{fig:colormnist_jitter}. (a) For long-tailed ColorMNIST, adding jitter makes solving the classification problem prohibitive, as color is required. Z2CNN and CECNN with jitter therefore perform no better than as the CECNN model with coset pooling. (b) For biased MNIST, performance decreases for small and improves for large $\sigma$, with CEConv still performing best.

\begin{figure}[ht!]
    \centering
    \includegraphics[width=0.5833\linewidth]{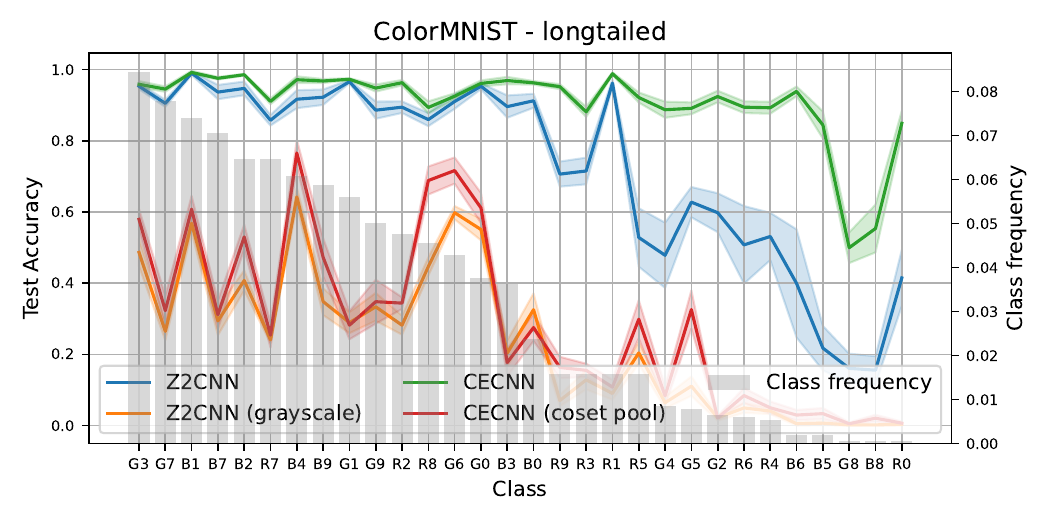}
    \caption{Per-class accuracy of various models trained with a loss function weighted by inverse class frequency. CECNN without coset pooling outperforms all other models, with no significant differences compared to an uniformly weighted loss function.}
    \label{fig:colormnist_longtailed_weighted}
\end{figure}

\paragraph{Long-tailed ColorMNIST with weighted loss}
We performed the longtailed ColorMNIST experiment both with a uniformly weighted loss and a loss where classes are weighted inversely to their frequency according to $w_i = \frac{N}{c * n_i}$, where $w_i$ denotes the weight for class $i$, $N$ the number of samples in the training set, $c$ the number of classes, and $n_i$ the number of samples for class $i$. The results are shown in \cref{fig:colormnist_longtailed_weighted}. We observed no significant difference between the two setups, with the CECNN without coset pooling outperforming the other models by a large margin in both.

\section{Classification experiments}
\subsection{Overview of all CE-ResNet configurations}
\label{app:cls}
\cref{tab:classification_std} shows an overview of the classification accuracies of all baselines and equivariant architectures. CEConv-x denotes the number of ResNet stages with CE convolutions with CEConv-4 (3 for CIFAR) being a fully equivariant ResNet. In nearly all cases, early equivariance is beneficial for improving classification accuracy on both the original as well as the hue shifted test sets. In case of the Flowers-102 dataset late equivariance seems to have a significant advantage, whereas for Caltech-101 and Stanford Cars the color equivariance bias does not seem to have much added value.

\cref{tab:classification_std} shows the classification results for all network architectures, trained with and without color jitter.

\begin{table}[!ht]
\setlength{\tabcolsep}{2pt}
\small
    \centering
    \rotatebox{90}{
    \begin{tabularx}{1.3\linewidth}{@{}lYYYYYYYc@{}}
        \toprule
        & \textbf{Caltech-101} & \textbf{CIFAR-10} & \textbf{CIFAR-100} & \textbf{Flowers-102} & \textbf{Oxford-IIIT Pet} & \textbf{Stanford Cars} & \textbf{STL-10} & \textbf{ImageNet} \\ \midrule
        \multicolumn{5}{@{}l}{\textit{Original test set}} \\ \midrule
        Baseline    & 71.61 $\pm$ 0.87 & 93.69 $\pm$ 0.16 & 71.28 $\pm$ 0.20 & 66.79 $\pm$ 0.89 & 69.87 $\pm$ 0.57 & 76.54 $\pm$ 0.10 & 83.80 $\pm$ 0.36 & 69.71 \\
        CIConv-W    & \textbf{72.85 $\pm$ 1.12} & 75.26 $\pm$ 0.57 & 38.81 $\pm$ 0.66 & 68.71 $\pm$ 0.29 & 61.53 $\pm$ 0.53 & \textbf{79.52 $\pm$ 0.42} & 80.71 $\pm$ 0.27 & 65.81 \\
        CEConv-1    & 71.59 $\pm$ 0.64 & \textbf{94.06 $\pm$ 0.09} & 71.82 $\pm$ 0.36 & 67.29 $\pm$ 0.57 & \textbf{70.47 $\pm$ 1.07} & 78.03 $\pm$ 0.29 & 84.34 $\pm$ 0.38 & \textbf{70.05} \\
        CEConv-2    & 71.50 $\pm$ 0.29 & 93.94 $\pm$ 0.07 & \textbf{72.20 $\pm$ 0.48} & 68.38 $\pm$ 0.55 & 70.34 $\pm$ 0.67 & 77.06 $\pm$ 0.38 & \textbf{84.50 $\pm$ 0.31} & 70.02 \\
        CEConv-3    & 70.45 $\pm$ 0.41 & 93.71 $\pm$ 0.26 & 71.37 $\pm$ 0.24 & \textbf{69.42 $\pm$ 0.58} & 68.92 $\pm$ 0.46 & 75.33 $\pm$ 0.66 & 83.61 $\pm$ 0.35 & 69.35 \\
        CEConv-4    & 70.16 $\pm$ 1.05 & -                & -                & 68.18 $\pm$ 0.45 & 70.24 $\pm$ 0.79 & 76.22 $\pm$ 0.19 & 84.24 $\pm$ 0.49 & 66.85 \\ \midrule
        Baseline + jitter   & 73.93 $\pm$ 0.73 & 93.03 $\pm$ 0.16 & 69.23 $\pm$ 0.44 & 68.75 $\pm$ 1.50 & 72.71 $\pm$ 0.67 & 80.59 $\pm$ 0.36 & 83.91 $\pm$ 0.38 & 69.37 \\
        CIConv-W + jitter   & \textbf{74.38 $\pm$ 0.43} & 77.49 $\pm$ 0.53 & 42.27 $\pm$ 0.56 & \textbf{75.05 $\pm$ 0.39} & 64.23 $\pm$ 0.51 & \textbf{81.56 $\pm$ 0.32} & 81.88 $\pm$ 0.24 & 65.95 \\
        CEConv-1 + jitter   & 73.43 $\pm$ 0.59 & \textbf{93.93 $\pm$ 0.16} & 71.08 $\pm$ 0.27 & 70.39 $\pm$ 0.81 & 72.44 $\pm$ 0.76 & 80.24 $\pm$ 0.51 & 84.31 $\pm$ 0.47 & 69.36 \\
        CEConv-2 + jitter   & 72.61 $\pm$ 0.95 & 93.86 $\pm$ 0.22 & \textbf{71.35 $\pm$ 0.20} & 71.72 $\pm$ 0.63 & 72.80 $\pm$ 0.87 & 80.32 $\pm$ 0.47 & \textbf{84.46 $\pm$ 0.39} & \textbf{69.42} \\
        CEConv-3 + jitter   & 73.21 $\pm$ 0.87 & 93.51 $\pm$ 0.10 & 71.12 $\pm$ 0.57 & 72.71 $\pm$ 0.23 & 72.55 $\pm$ 0.67 & 79.62 $\pm$ 0.54 & 84.08 $\pm$ 0.44 & 69.10 \\
        CEConv-4 + jitter   & 73.58 $\pm$ 0.68 & -                & -                & 74.17 $\pm$ 0.49 & \textbf{73.28 $\pm$ 0.63} & 79.79 $\pm$ 0.37 & 84.16 $\pm$ 0.10 & 65.57 \\ \midrule
        Baseline + AugMix  & \textbf{71.92 $\pm$ 0.95} & 94.13 $\pm$ 0.22 & \textbf{72.64 $\pm$ 0.27} & 75.49 $\pm$ 0.24 & \textbf{76.02 $\pm$ 0.51} & \textbf{82.32 $\pm$ 0.07} & 84.99 $\pm$ 0.24 & - \\
        CEConv + AugMix    & 70.74 $\pm$ 1.12 & \textbf{94.22 $\pm$ 0.16} & 72.48 $\pm$ 0.18 & \textbf{78.10 $\pm$ 0.50} & 75.90 $\pm$ 0.22 & 80.81 $\pm$ 0.27 & \textbf{85.46 $\pm$ 0.30} & - \\ \midrule
        \multicolumn{5}{@{}l}{\textit{Hue-shifted test set}} \\ \midrule
        Baseline    & 51.14 $\pm$ 0.71 & 85.26 $\pm$ 0.56 & 47.01 $\pm$ 0.38 & 13.41 $\pm$ 0.34 & 37.56 $\pm$ 0.76 & 55.59 $\pm$ 0.74 & 67.60 $\pm$ 0.56 & 54.72 \\
        CIConv-W    & \textbf{71.92 $\pm$ 1.11} & 74.88 $\pm$ 0.54 & 37.09 $\pm$ 0.74 & \textbf{59.03 $\pm$ 0.62} & \textbf{60.54 $\pm$ 0.46} & \textbf{78.71 $\pm$ 0.33} & \textbf{79.92 $\pm$ 0.25} & \textbf{64.62} \\
        CEConv-1    & 65.60 $\pm$ 0.47 & \textbf{91.93 $\pm$ 0.14} & \textbf{63.37 $\pm$ 0.17} & 32.88 $\pm$ 0.83 & 52.97 $\pm$ 1.00 & 70.08 $\pm$ 0.21 & 78.83 $\pm$ 0.43 & 63.02 \\
        CEConv-2    & 64.51 $\pm$ 0.64 & 91.43 $\pm$ 0.18 & 62.11 $\pm$ 0.43 & 33.32 $\pm$ 0.55 & 51.14 $\pm$ 0.95 & 68.17 $\pm$ 0.86 & 77.80 $\pm$ 0.58 & 62.26 \\
        CEConv-3    & 62.22 $\pm$ 0.99 & 90.90 $\pm$ 0.25 & 59.04 $\pm$ 0.45 & 33.76 $\pm$ 0.38 & 49.45 $\pm$ 0.65 & 65.82 $\pm$ 1.34 & 76.23 $\pm$ 0.37 & 60.95 \\
        CEConv-4    & 62.17 $\pm$ 1.01 & -                & -                & 33.33 $\pm$ 0.38 & 54.02 $\pm$ 1.34 & 67.16 $\pm$ 0.58 & 78.25 $\pm$ 0.52 & 56.90 \\ \midrule
        Baseline + jitter   & 73.61 $\pm$ 0.60 & 92.91 $\pm$ 0.17 & 69.12 $\pm$ 0.47 & 68.44 $\pm$ 1.60 & 72.31 $\pm$ 0.49 & 80.65 $\pm$ 0.36 & 83.71 $\pm$ 0.35 & 67.10 \\
        CIConv-W + jitter   & \textbf{74.40 $\pm$ 0.55} & 77.28 $\pm$ 0.54 & 42.30 $\pm$ 0.48 & \textbf{75.66 $\pm$ 0.27} & 63.93 $\pm$ 0.42 & \textbf{81.44 $\pm$ 0.26} & 81.54 $\pm$ 0.21 & 65.03 \\
        CEConv-1 + jitter   & 73.34 $\pm$ 0.96 & \textbf{93.86 $\pm$ 0.20} & 70.98 $\pm$ 0.22 & 69.98 $\pm$ 0.79 & 72.34 $\pm$ 0.58 & 80.18 $\pm$ 0.50 & 84.29 $\pm$ 0.50 & \textbf{68.85} \\
        CEConv-2 + jitter   & 73.03 $\pm$ 0.97 & 93.80 $\pm$ 0.14 & \textbf{71.33 $\pm$ 0.19} & 71.44 $\pm$ 0.57 & 72.58 $\pm$ 0.86 & 80.28 $\pm$ 0.52 & \textbf{84.31 $\pm$ 0.34} & 68.74 \\
        CEConv-3 + jitter   & 73.26 $\pm$ 0.74 & 93.39 $\pm$ 0.08 & 71.06 $\pm$ 0.53 & 72.47 $\pm$ 0.20 & 72.32 $\pm$ 0.64 & 79.62 $\pm$ 0.54 & 84.00 $\pm$ 0.33 & 68.03 \\
        CEConv-4 + jitter   & 73.57 $\pm$ 0.75 & -                & -                & 73.86 $\pm$ 0.39 & \textbf{72.94 $\pm$ 0.56} & 79.79 $\pm$ 0.34 & 84.02 $\pm$ 0.14 & 64.52 \\ \midrule
        Baseline + AugMix   & 51.82 $\pm$ 0.60 & 88.03 $\pm$ 0.26 & 51.39 $\pm$ 0.19 & 15.99 $\pm$ 0.28 & 48.04 $\pm$ 0.74 & 68.69 $\pm$ 0.73 & 72.19 $\pm$ 0.45 & - \\
        CEConv + AugMix     & \textbf{62.29 $\pm$ 0.97} & \textbf{91.68 $\pm$ 0.21} & \textbf{60.75 $\pm$ 0.24} & \textbf{41.43 $\pm$ 0.97} & \textbf{62.27 $\pm$ 0.81} & \textbf{73.59 $\pm$ 0.30} & \textbf{80.17 $\pm$ 0.15} & - \\
    \end{tabularx}}
    \caption{Classification accuracy on various datasets. CEConv-$s$ denotes a ResNet with $s$ color equivariant stages. We report results for models trained with and without color jitter augmentation. (Hybrid) color equivariant networks improve performance over the baseline model on both the original as well as the hue-shifted test set.}
    \label{tab:classification_std}
\end{table}

\newpage
\subsection{Test-time hue shift plots}
\label{app:test_hue_shift}
\begin{figure*}[ht!]
\centering
\begin{subfigure}{0.5\textwidth}
\centering
\includegraphics[width=\linewidth]{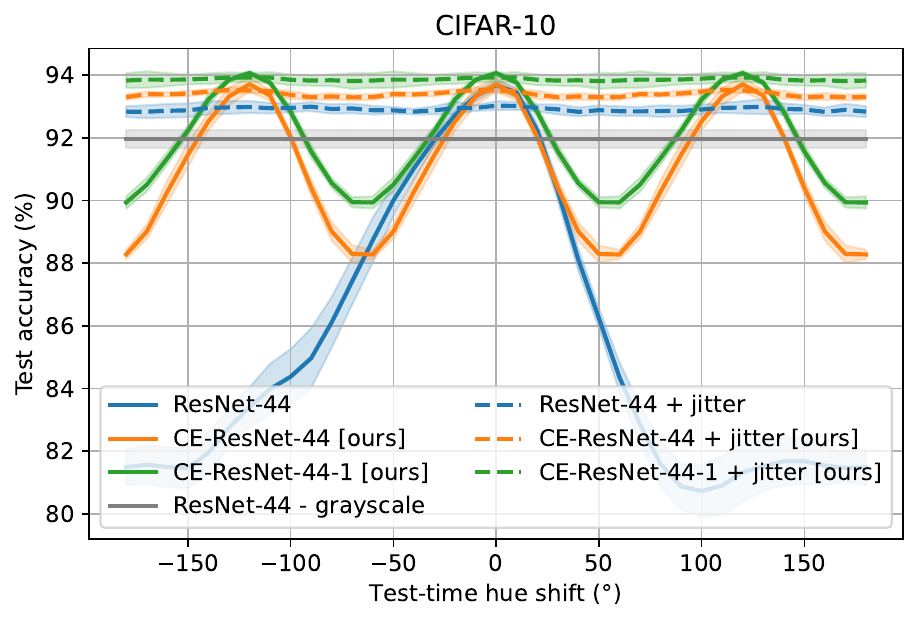}
\end{subfigure}%
\begin{subfigure}{0.5\textwidth}
\centering
\includegraphics[width=\linewidth]{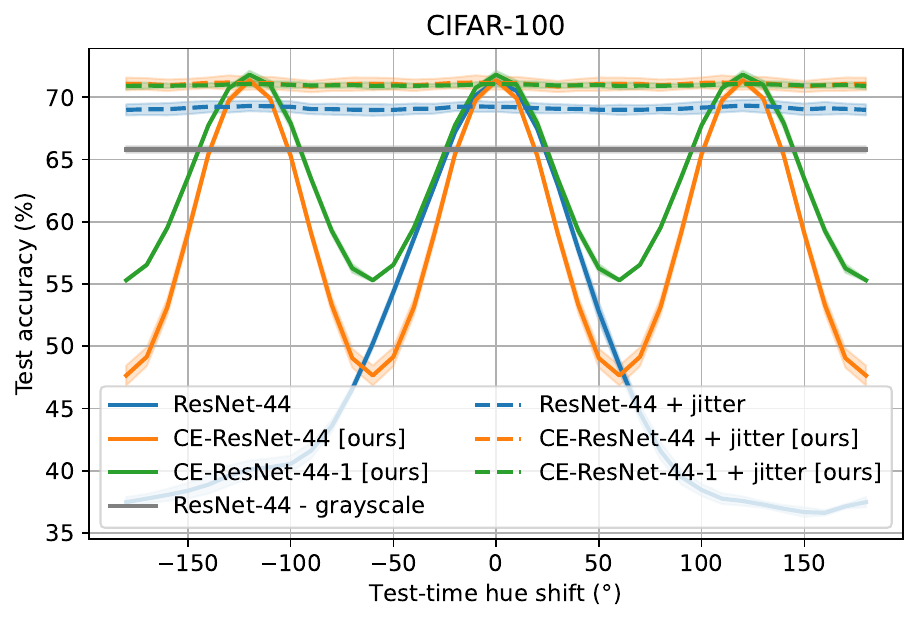}
\end{subfigure}
\begin{subfigure}{0.5\textwidth}
\centering
\includegraphics[width=\linewidth]{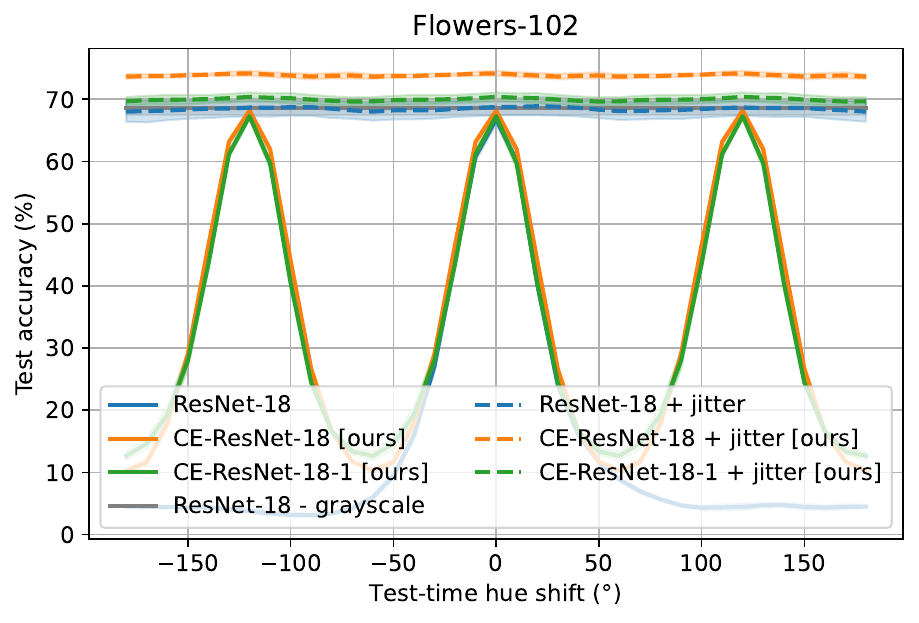}
\end{subfigure}%
\begin{subfigure}{0.5\textwidth}
\centering
\includegraphics[width=\linewidth]{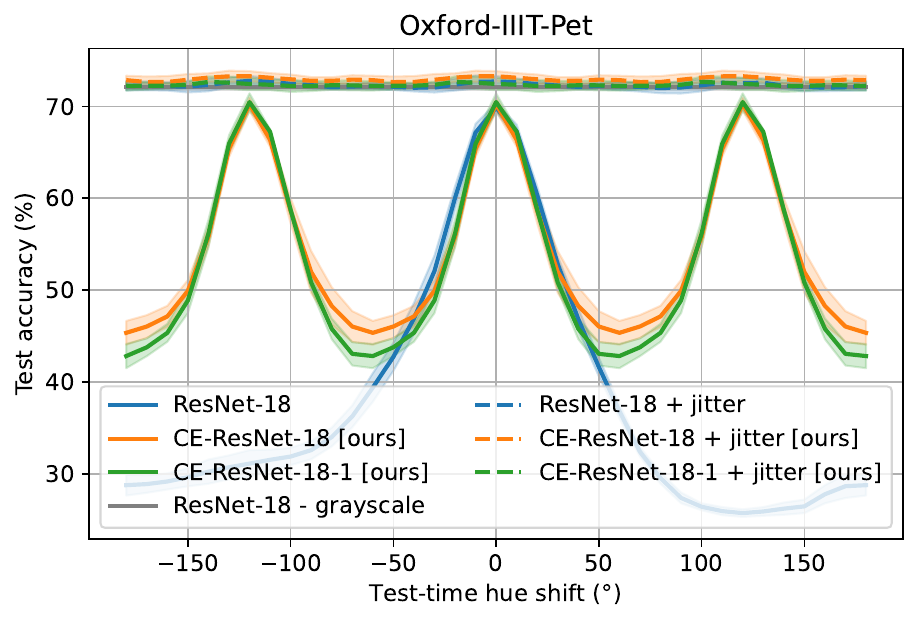}
\end{subfigure}
\begin{subfigure}{0.5\textwidth}
\centering
\includegraphics[width=\linewidth]{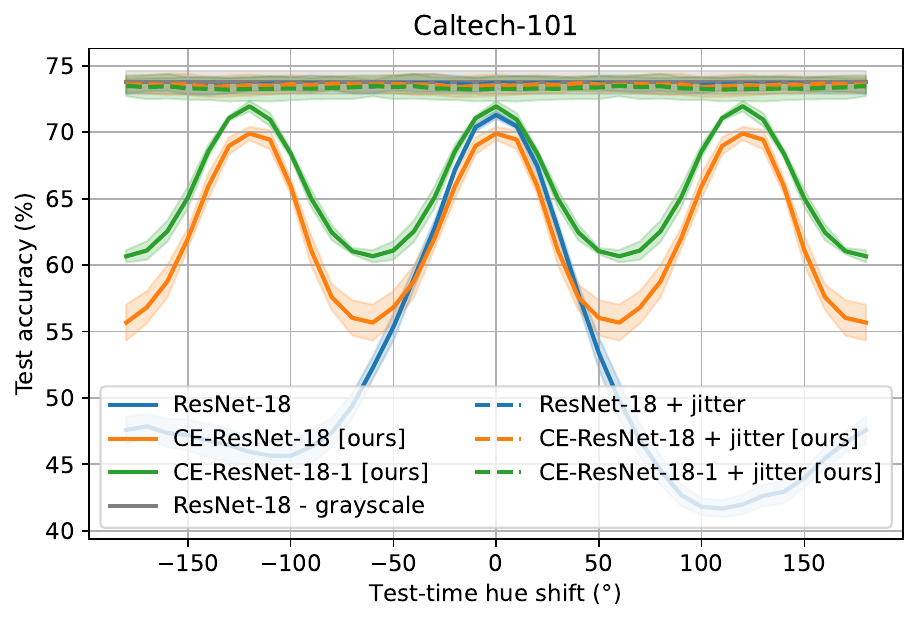}
\end{subfigure}%
\begin{subfigure}{0.5\textwidth}
\centering
\includegraphics[width=\linewidth]{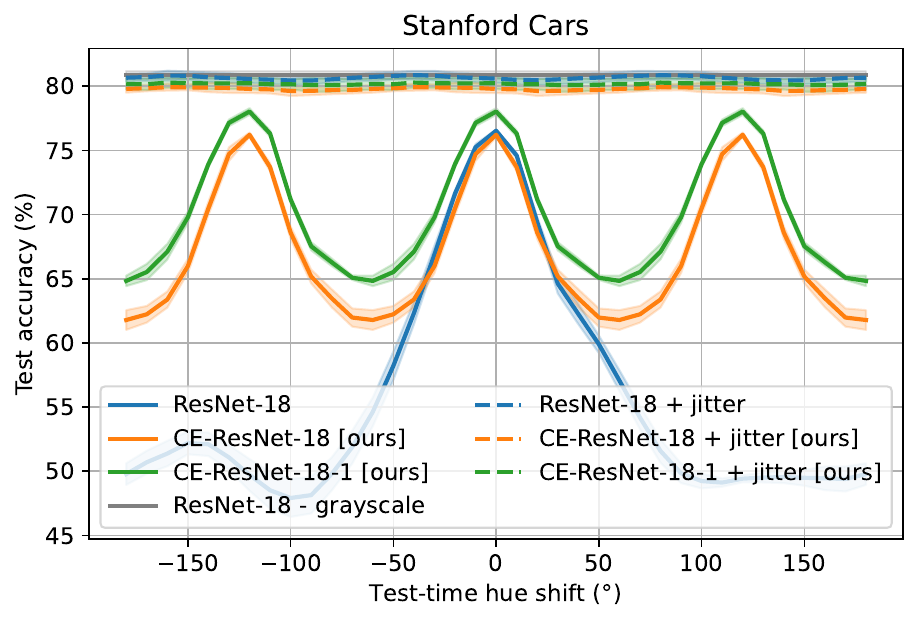}
\end{subfigure}
\begin{subfigure}{0.5\textwidth}
\centering
\includegraphics[width=\linewidth]{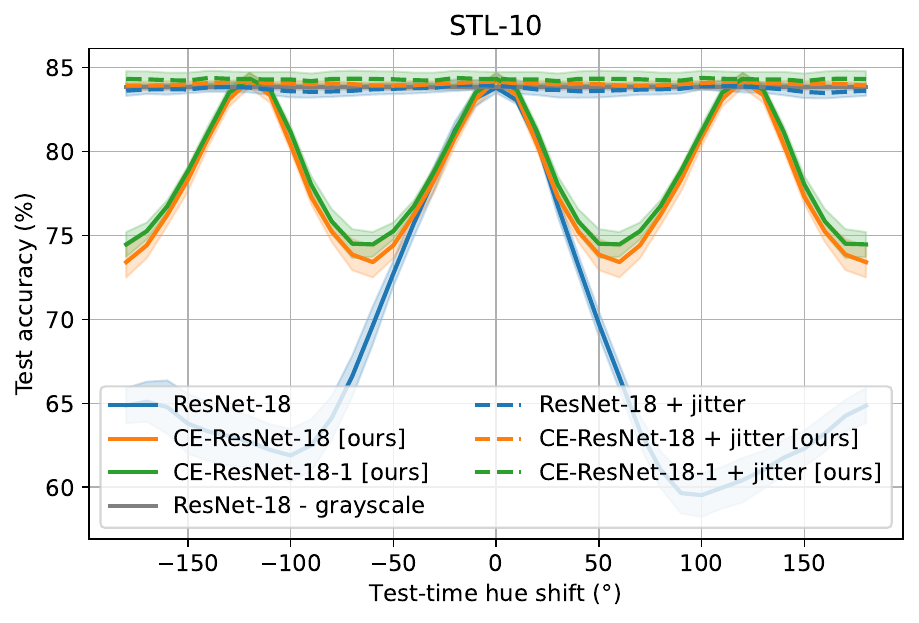}
\end{subfigure}%
\begin{subfigure}{0.5\textwidth}
\centering
\includegraphics[width=\linewidth]{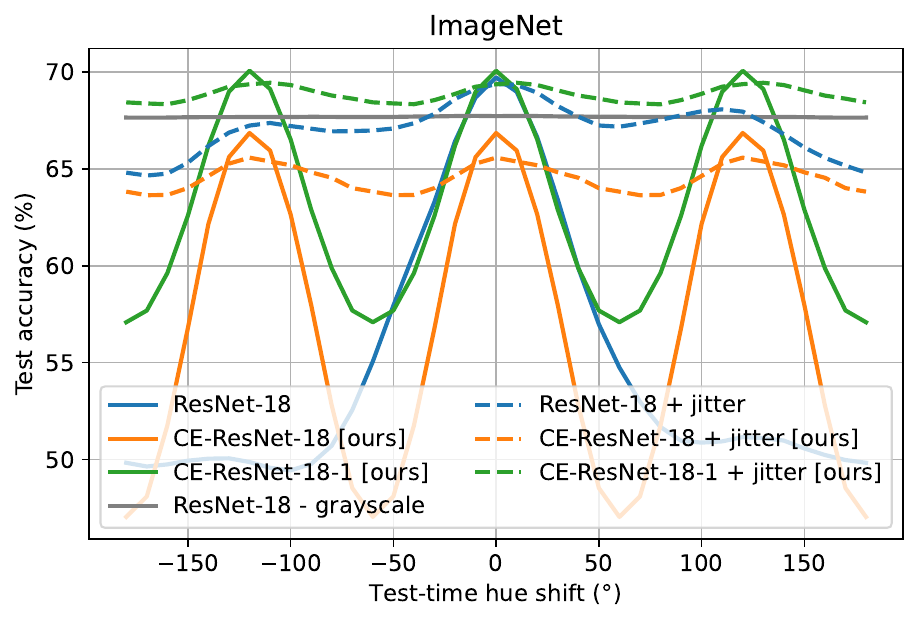}
\end{subfigure}
\caption{Test accuracy on various classification datasets under a test time hue shift.}
\label{fig:classification_complete}
\end{figure*}

\cref{fig:classification_complete} shows the test accuracies under a test time hue shift on all datasets in the paper. Each figure includes a regular ResNet, a color equivariant ResNet-x (CE-ResNet-x) and a ResNet-x with color equivariant convolutions in the first ResNet stage (CE-ResNet-x-1), trained with and without color jitter augmentation.  Finally, the plot shows the accuracy of a ResNet-x trained on grayscale inputs. CEConv improves robustness to test-time hue shifts on all datasets.

\newpage

\subsection{CE-ResNet configurations}
\label{app:configs}
The configurations of the color equivariant ResNet with three hue rotations, as used in the classification experiment in \cref{sec:image_classification}, are shown in \cref{tab:resnet_configs}. CE stages 0 denotes a regular ResNet.

\begin{table}[ht!]
\centering
\begin{tabularx}{0.95\linewidth}{@{}lYYYY@{}}
\toprule
\textbf{Model} & \textbf{CE stages} & \textbf{Width} & \textbf{Parameters (M)} & \textbf{MACs (G)} \\ \midrule
\multirow{5}{*}{ResNet-18}
 & 0 & 64 & 11.69 & 3.59 \\
 & 1 & 63 & 11.38 & 5.66 \\
 & 2 & 63 & 11.57 & 7.37 \\
 & 3 & 61 & 11.54 & 8.80 \\
 & 4 & 55 & 11.79 & 10.32 \\ \midrule
\multirow{4}{*}{ResNet-44}
 & 0 & 32 & 2.64 & 0.78 \\
 & 1 & 31 & 2.51 & 1.23 \\
 & 2 & 30 & 2.50 & 1.63 \\
 & 3 & 27 & 2.60 & 1.83 \\ \bottomrule
\end{tabularx}
\caption{Color equivariant ResNet configurations.}
\label{tab:resnet_configs}
\end{table}

\subsection{Neuron Feature visualizations}
\label{app:nf_vis}
\begin{figure}[ht!]
    \centering
    \includegraphics[width=\linewidth]{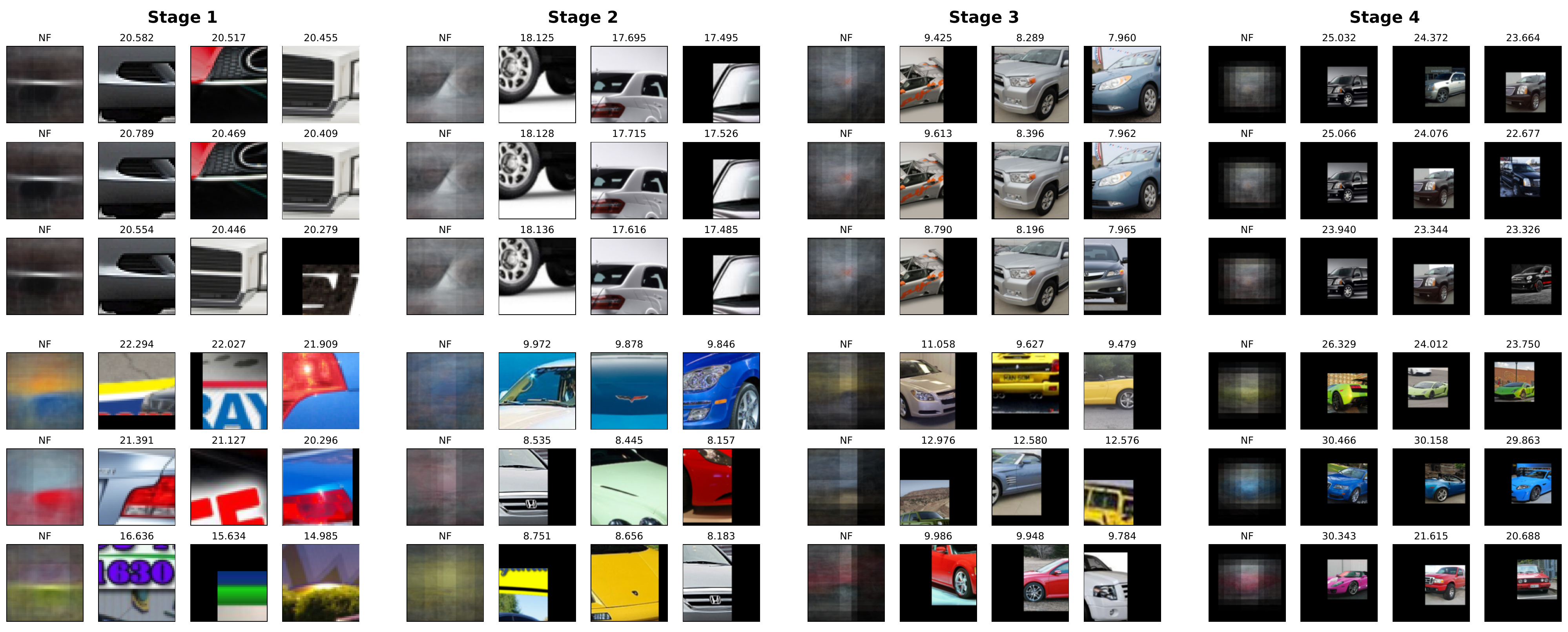}
    \caption{Neuron Feature~\cite{rafegas2018color} (NF) visualization with top-3 patches of two neurons at different stages in a CE-ResNet18 trained on Stanford Cars. Rows represent different rotations of the same filter.}
    \label{fig:neuronfeature_sup}
\end{figure}

\cref{fig:neuronfeature_sup} shows the Neuron Feature~\cite{rafegas2018color} (NF) visualization with top-3 patches of two neurons at different stages in a CE-ResNet18 trained on Stanford Cars. As expected, each row of a NF activates on the same shape in a different color. We show neurons that are insensitive to color (top row) and neurons that are sensitive to color (bottom row).

\section{Ablation studies}
\label{app:ablation}

\paragraph{Strength of color jitter augmentations}

\begin{figure}[ht!]
\begin{subfigure}{0.5\textwidth}
\centering
\includegraphics[width=\linewidth]{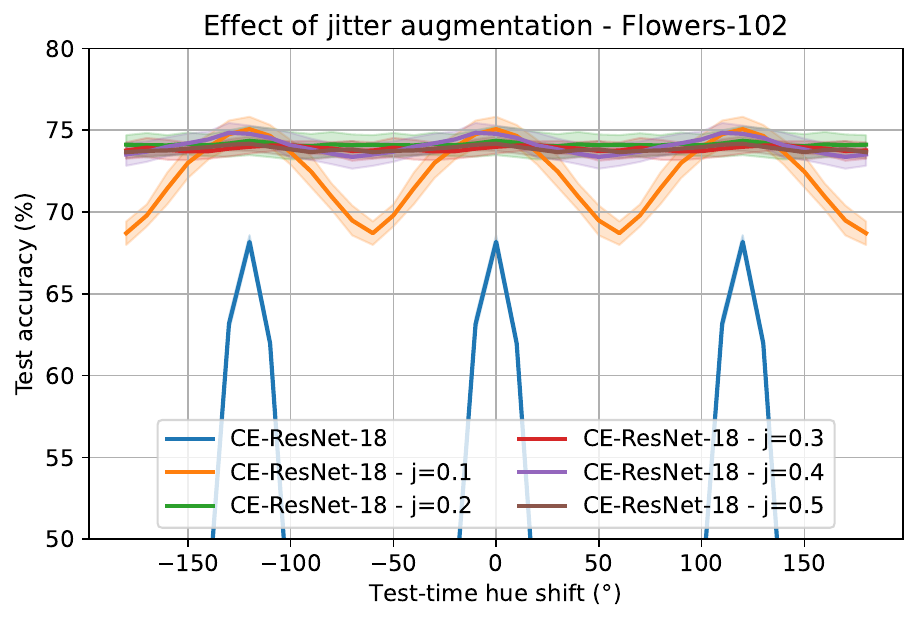}
\caption{}
\end{subfigure}%
\begin{subfigure}{0.5\textwidth}
\centering
\includegraphics[width=\linewidth]{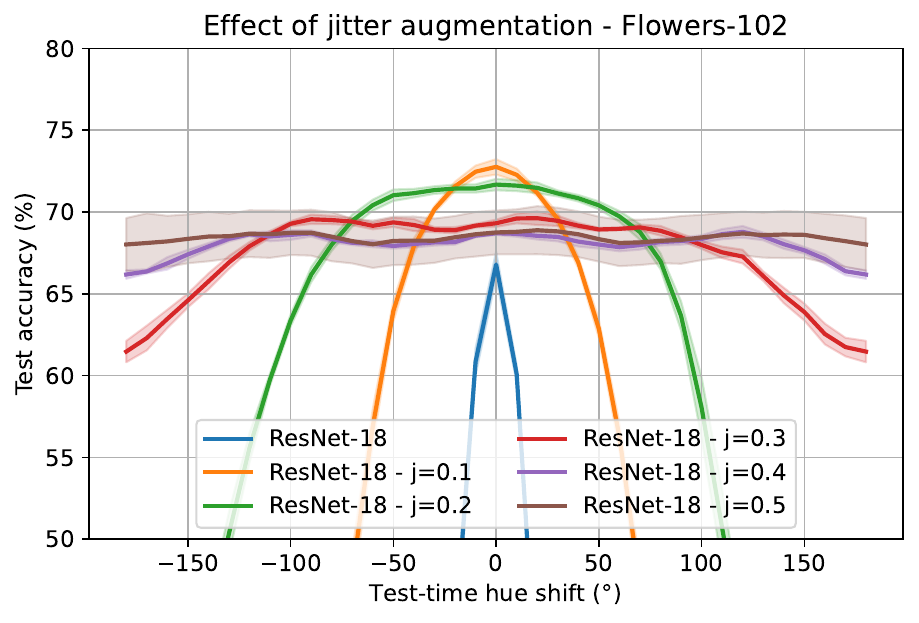}
\caption{}
\end{subfigure}
\caption{Effect of hue jitter augmentation on a color equivariant (a) and a regular (b) ResNet-18.}
\label{fig:ablations_jitter}
\end{figure}

\cref{fig:ablations_jitter} shows the effect of hue jitter augmentation during training on both a color equivariant ResNet-18 with 3 rotations (a) and a regular ResNet-18 (b) trained on Flowers-102. All runs have been repeated 3 times and the mean performance is reported. As expected, the color equivariant network (a) without jitter augmentation is equivariant to rotations of multiples of 120 degrees, but performance quickly degrades. Applying slight (0.1) hue jitter during training both helps in an absolute sense, increasing performance over all rotations, and makes the network more robust to hue changes as shown by the increasing width of the peaks. Further increasing the strength of the augmentation results in a uniform performance over all hue shifts, indicated by the flat lines. There appears to be no significant difference for jitter strength $> 0.2$.  In comparison, the regular ResNet (b) trained without hue augmentation shows a single peak around 0 degrees, which increases in width when applying more severe augmentation. Note that the increase in absolute performance is smaller compared to the color equivariant network. The reason for this is that the equivariant architecture only requires augmentation "between" the discrete rotations to which it is already robust, as opposed to the full scale of hue shifts for the baseline architecture. Augmentation and equivariance thus exhibit a remarkable synergistic interaction.

\paragraph{Group coset pooling}

\begin{figure}[ht!]
    \centering
    \includegraphics[width=0.5\linewidth]{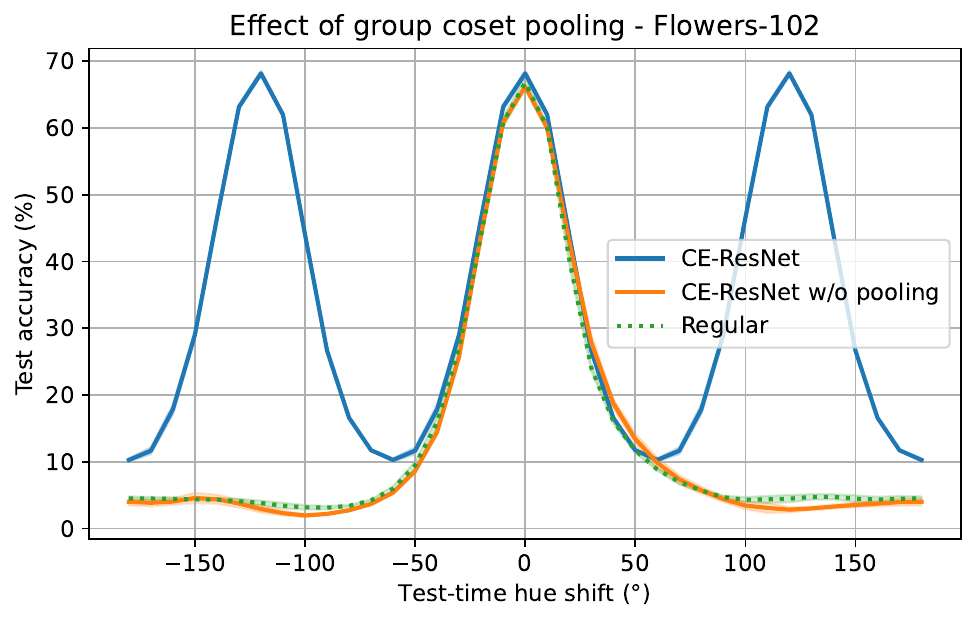}
    \caption{CE-ResNet without group coset pooling behaves similarly to a regular ResNet (average over 5 runs).}
    \label{fig:ablation_cosetpool}
\end{figure}

We have removed the group coset pooling operation by flattening the feature map group dimension into the channel dimension in the penultimate layer, before applying the final classification layer. As shown in \cref{fig:ablation_cosetpool}, the model without pooling layer is no longer invariant to hue shifts and behaves identically to the baseline model.

\begin{figure}[ht!]
    \begin{subfigure}{0.5\textwidth}
        \centering
        \includegraphics[width=\linewidth]{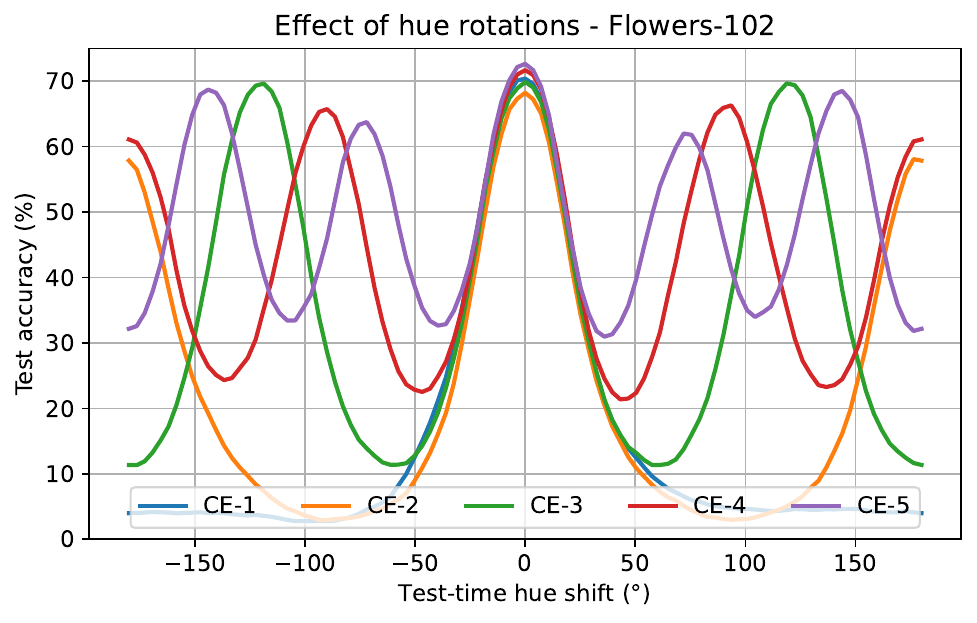}
        \caption{Test-time hue shift with reprojection.}
    \end{subfigure}%
    \begin{subfigure}{0.5\textwidth}
        \centering
        \includegraphics[width=\linewidth]{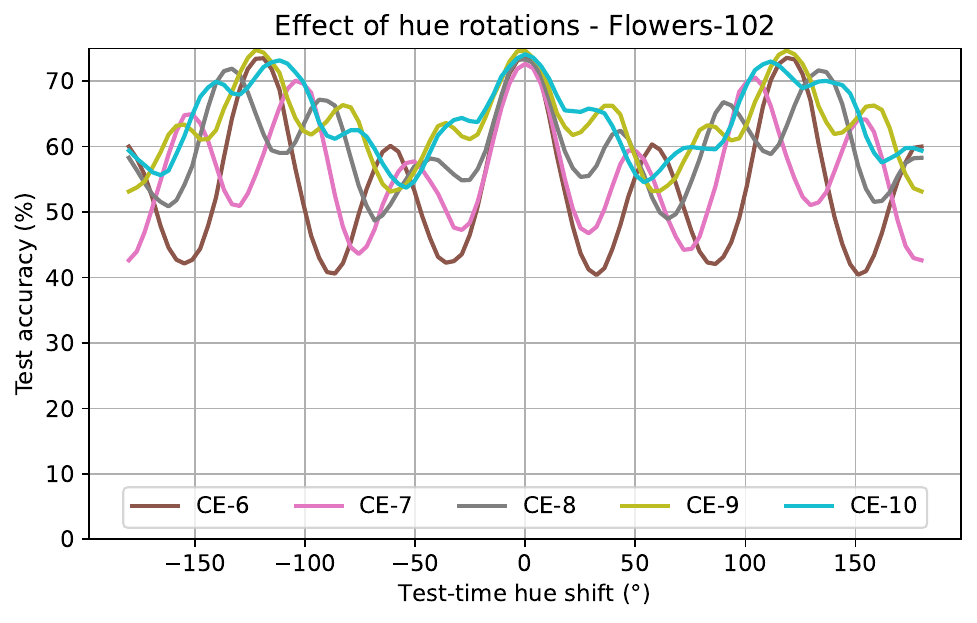}
        \caption{Test-time hue shift with reprojection.}
    \end{subfigure}
    \begin{subfigure}{0.5\textwidth}
        \centering
        \includegraphics[width=\linewidth]{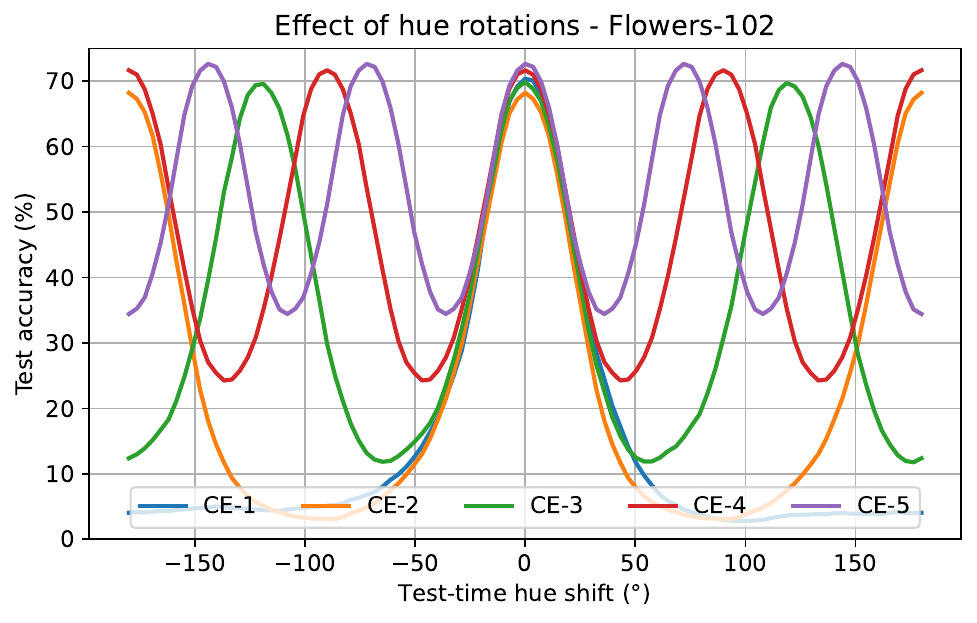}
        \caption{Test-time hue shift without reprojection.}
    \end{subfigure}%
    \begin{subfigure}{0.5\textwidth}
        \centering
        \includegraphics[width=\linewidth]{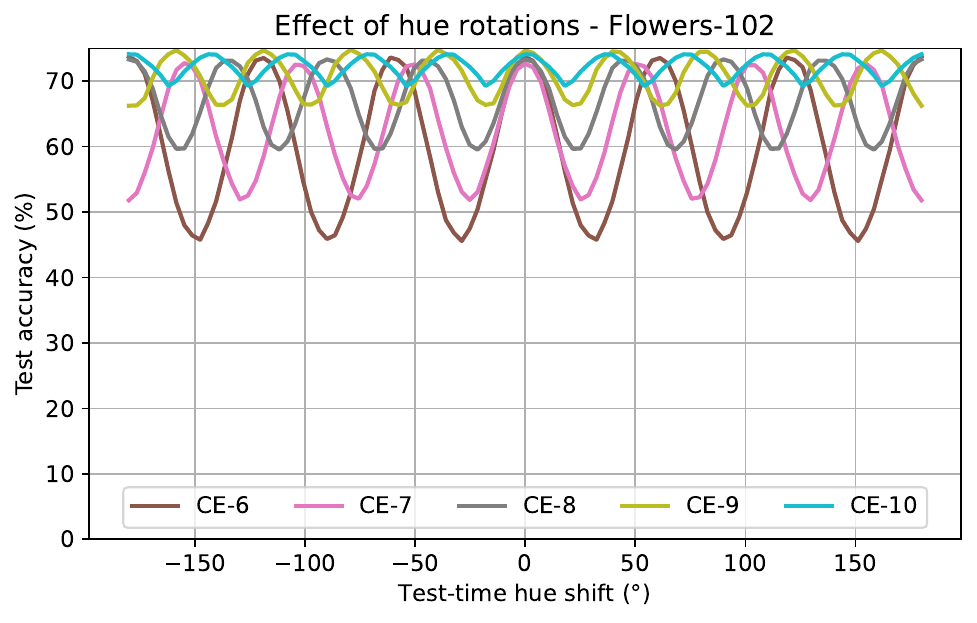}
        \caption{Test-time hue shift without reprojection.}
    \end{subfigure}
    \caption{The effect of the number of hue rotations in color equivariant convolutions on downstream performance. More rotations increases robustness to test-time hue shifts. Note that in (a-b) the peaks are not of equal height due to clipping effects near the boundaries of the RGB cube. This artifact disappears when the test-time hue shift is also applied without reprojection, resulting in peaks of equal height (c-d).}
    \label{fig:ablations_rotations}
\end{figure}

\paragraph{Number of color rotations} We investigate the effect of the number of hue rotations in color equivariant convolutions by training CE-ResNets with 2-10 rotations on Flowers-102. \cref{fig:ablations_rotations} shows the test accuracies for rotations 1-5 (a) and 6-10 (b), respectively. Note that, for this particular dataset, more hue rotations not only lead to better robustness to test-time hue shifts, but also to better absolute performance. However, there is a trade-off between number of rotations and model capacity, as increasing the number of rotations increases the number of parameters in the model, and the model width needs to be scaled down to keep the number of parameters equal. Both the optimal number of color rotations and network width therefore depend on the amount of color vs. the complexity of the data, and therefore both need to be carefully calibrated per dataset.

As expected, the number of peaks increases with the number of hue rotations, though interestingly, the peaks do vary in height. This is an artifact due the way test-time hue shifts are applied to the input images. When RGB pixels are rotated about the [1,1,1] diagonal, values near the borders of the RGB cube tend to fall outside the cube and subsequently need to be reprojected. This reprojection is not modeled by the filter transformations in the CEConv layers, and subsequently causes a discrepancy between the filter and the image transformations.  Indeed, when the test-time hue shift is instead implemented through a rotation in RGB space without reprojecting into the cube, this artifact disappears and all peaks are of equal height, as shown in \cref{fig:ablations_rotations} (c-d). Note that rotations of multiples of 120 degrees always end up within the RGB cube, which is why this artifact does never occur at -120, 0 and 120 degrees. Future work should further investigate the extent to which this discrepancy is problematic in practice, and look into alternative solutions.

\end{document}